\documentclass[11pt,a4paper]{article}
\usepackage[nohyperref]{emnlp-ijcnlp-2019}

\usepackage{latexsym}

\usepackage{times}
\usepackage{relsize}
\usepackage[T1]{fontenc}
\usepackage[scaled=0.90]{inconsolata}
\usepackage[latin1]{inputenc}
\usepackage[english]{babel}

\usepackage{epsfig}
\usepackage{graphicx}
\usepackage{wrapfig}
\usepackage[belowskip=0pt,aboveskip=0pt,font=small]{caption}
\usepackage[belowskip=0pt,aboveskip=0pt,font=small]{subcaption}
\usepackage{amsmath, amsthm, amssymb, mathabx}
\usepackage{textcomp}
\usepackage{stmaryrd}
\SetSymbolFont{stmry}{bold}{U}{stmry}{m}{n}
\usepackage{upgreek}
\usepackage{bm}
\usepackage{cases}
\usepackage{mathtools}
\usepackage{arydshln}
\usepackage{multirow}

\definecolor{bluebell}{RGB}{52,31,151}
\definecolor{amour}{RGB}{238,82,83}
\usepackage[pagebackref=true,breaklinks=true,colorlinks,bookmarks=false,citecolor=bluebell,linkcolor=amour]{hyperref}

\usepackage{multirow}
\usepackage{rotating}
\usepackage{booktabs}

\usepackage{enumitem}
\usepackage[olditem,oldenum]{paralist}

\usepackage{alltt}
\usepackage{listings}

\usepackage{url}
\usepackage{xspace}
\usepackage{comment}
\usepackage{color}
\usepackage{afterpage}
\usepackage{framed}
\usepackage{fancybox}
\usepackage{cuted}
\usepackage{caption}

\usepackage{quoting}
\quotingsetup{vskip=0pt}
\usepackage{epigraph}

\usepackage{mysymbols}

\belowdisplayskip \abovedisplayskip

\newlength{\sectionReduceTop}
\newlength{\sectionReduceBot}
\newlength{\subsectionReduceTop}
\newlength{\subsectionReduceBot}
\newlength{\abstractReduceTop}
\newlength{\abstractReduceBot}
\newlength{\captionReduceTop}
\newlength{\captionReduceBot}
\newlength{\subsubsectionReduceTop}
\newlength{\subsubsectionReduceBot}

\newlength{\eqnReduceTop}
\newlength{\eqnReduceBot}

\newlength{\horSkip}
\newlength{\verSkip}

\newlength{\figureHeight}
\definecolor{belize}{RGB}{41, 128, 185}

\newcommand{\Qbot}{\textsc{Q-bot}\xspace}
\newcommand{\Abot}{\textsc{A-bot}\xspace}

\newcommand{\figref}[1]{Fig.~\ref{#1}}
\newcommand{\secref}[1]{Section \ref{#1}}

\aclfinalcopy %

\title{Improving Generative Visual Dialog by Answering Diverse Questions}

\author{Vishvak Murahari$^1$ \quad Prithvijit Chattopadhyay$^1$ \\[0.05in]
  \textbf{Dhruv Batra$^{1,2}$ \quad Devi Parikh$^{1,2}$ \quad Abhishek Das$^1$} \\[0.2in]
  $^1$Georgia Tech \quad
  $^2$Facebook AI Research}

\date{}

\begin{document}
\maketitle
\begin{abstract}
  Prior work on training generative Visual Dialog models with reinforcement learning
  \cite{visdial_rl} has explored a \Qbot-\Abot image-guessing game and shown that
  this `self-talk' approach can lead to improved performance at the downstream dialog-conditioned
  image-guessing task. However, this improvement saturates and starts degrading after
  a few rounds of interaction, and does not lead to a better Visual Dialog model.
  We find that this is due in part to repeated interactions between \Qbot and \Abot during
  self-talk, which are not informative with respect to the image. To improve this,
  we devise a simple auxiliary objective that incentivizes \Qbot to ask diverse
  questions, thus reducing repetitions and in turn enabling \Abot to explore a
  larger state space during RL~\ie be exposed to more visual concepts to talk about, and
  varied questions to answer. We evaluate our approach via a host of automatic
  metrics and human studies, and demonstrate that it leads to
  better dialog,~\ie dialog that is more diverse (\ie less repetitive),
  consistent (\ie has fewer conflicting exchanges), fluent (\ie more human-like),
  and detailed, while still being comparably image-relevant as prior work and ablations.
  Our code is publicly available at \href{https://github.com/vmurahari3/visdial-diversity}{{\tt github.com/vmurahari3/visdial-diversity}}.
\end{abstract}

\section{Introduction}
\label{sec:intro}

\begin{figure*}[t!]
    \begin{minipage}[b]{0.7\textwidth}
        \begin{subfigure}{\textwidth}
            \centering
            \includegraphics[width=0.98\textwidth]{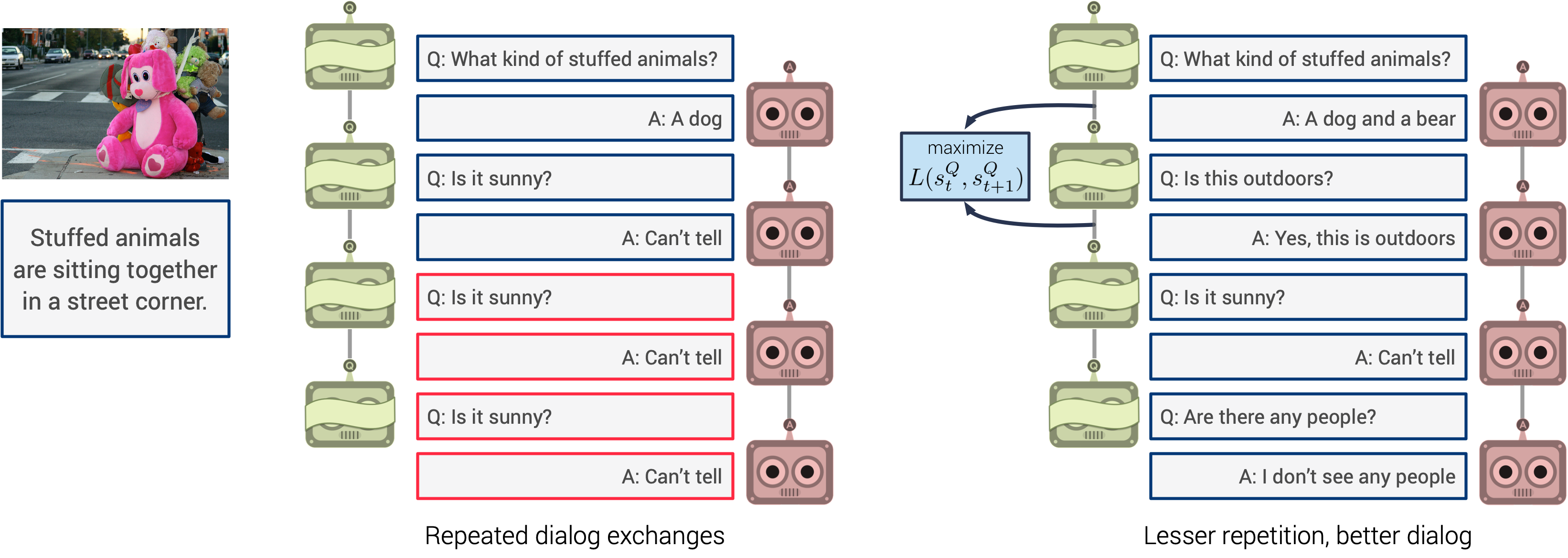}
            \vspace{5pt}
            \caption{{\scriptsize
                \textbf{Left}. Prior work on training generative Visual Dialog models with RL
                on an image-guessing task between \Qbot and \Abot~\cite{visdial_rl} leads to repetitive dialog.
                \textbf{Right}. We devise an auxiliary objective that incentivizes \Qbot to ask diverse
                questions, thus reducing repetitions and enabling \Abot to be exposed to more
                varied questions during RL, overall leading to better dialog, as measured by automatic metrics and human studies.
            }}
            \label{fig:teaser}
        \end{subfigure}
    \end{minipage}\hfill
    \begin{minipage}[b]{0.27\textwidth}
        \begin{subfigure}{\textwidth}
            \centering
            \includegraphics[width=0.95\textwidth]{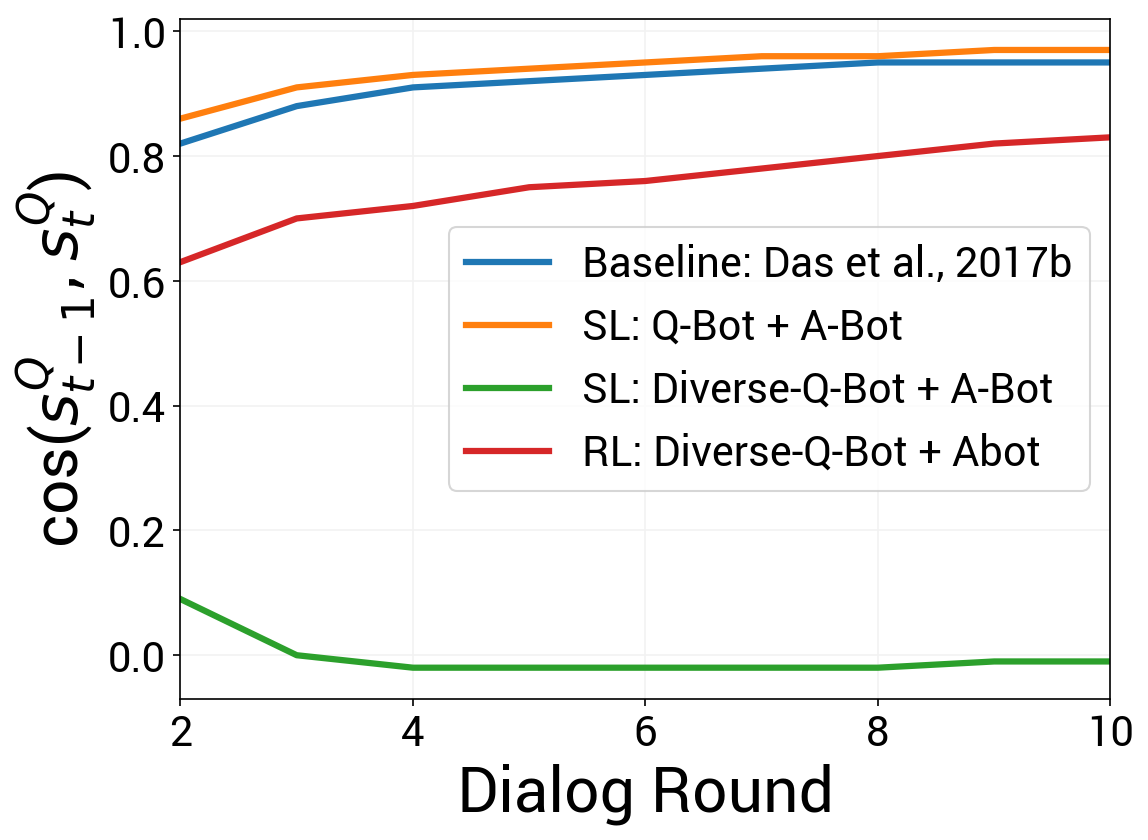}
            \vspace{21pt}
            \caption{{\scriptsize Cosine similarity of successive dialog state embeddings
                within \Qbot. Prior work~\cite{visdial_rl} has high
                similarity. Our approach explicitly
                minimizes this similarity leading to more diverse dialog.}}
            \label{fig:similar_embeddings}
        \end{subfigure}
    \end{minipage}\hfill
    \vspace{5pt}
    \caption{}
    \vspace{-10pt}
\end{figure*}

Our goal is to build agents that can \emph{see} and
\emph{talk} \ie agents that can perceive the visual world and communicate this
understanding in natural language conversations in English.
To this end, \citet{visdial,vries_cvpr17} proposed the task of Visual Dialog --
given an image, dialog history consisting of a sequence of question-answer pairs,
and a follow-up question about the image, predict a free-form natural language
answer to the question -- along with a dataset and evaluation
metrics\footnote{\href{https://visualdialog.org}{\tt{visualdialog.org}}, \href{https://guesswhat.ai}{\tt{guesswhat.ai}}}.

Posing Visual Dialog as a supervised learning problem is unnatural.
This is because at every round of dialog, the agent's answer prediction is thrown away
and it gets access to ground-truth dialog history, thus disabling it from steering
conversations during training. This leads to compounding errors over long-range
sequences at test time, a problem also common in training recurrent neural networks
for language modeling~\cite{bengio_nips15,ross_aistats11,ranzato_iclr16}.

To overcome this, \citet{visdial_rl} devised a goal-driven approach for training
Visual Dialog agents with deep reinforcement learning.
This is formulated as a game between two agents -- \Qbot (that asks questions) and \Abot
(that answers questions). \Qbot is shown a one-line description of an unseen image
that \Abot has access to and \Qbot is allowed to ask questions (in natural language)
to \Abot for a fixed number of rounds and simultaneously make predictions of the
unseen image. Both agents are rewarded for \Qbot's image-guessing performance
and trained with REINFORCE~\cite{williams1992simple} to optimize this reward.
Thus, there is incentive for \Qbot to ask questions informative of the hidden image,
and for \Abot to provide meaningful answers.

\begin{figure*}[t!]
    \begin{minipage}[b]{0.26\textwidth}
        \begin{subfigure}{\textwidth}
            \centering
            \includegraphics[width=0.9\textwidth]{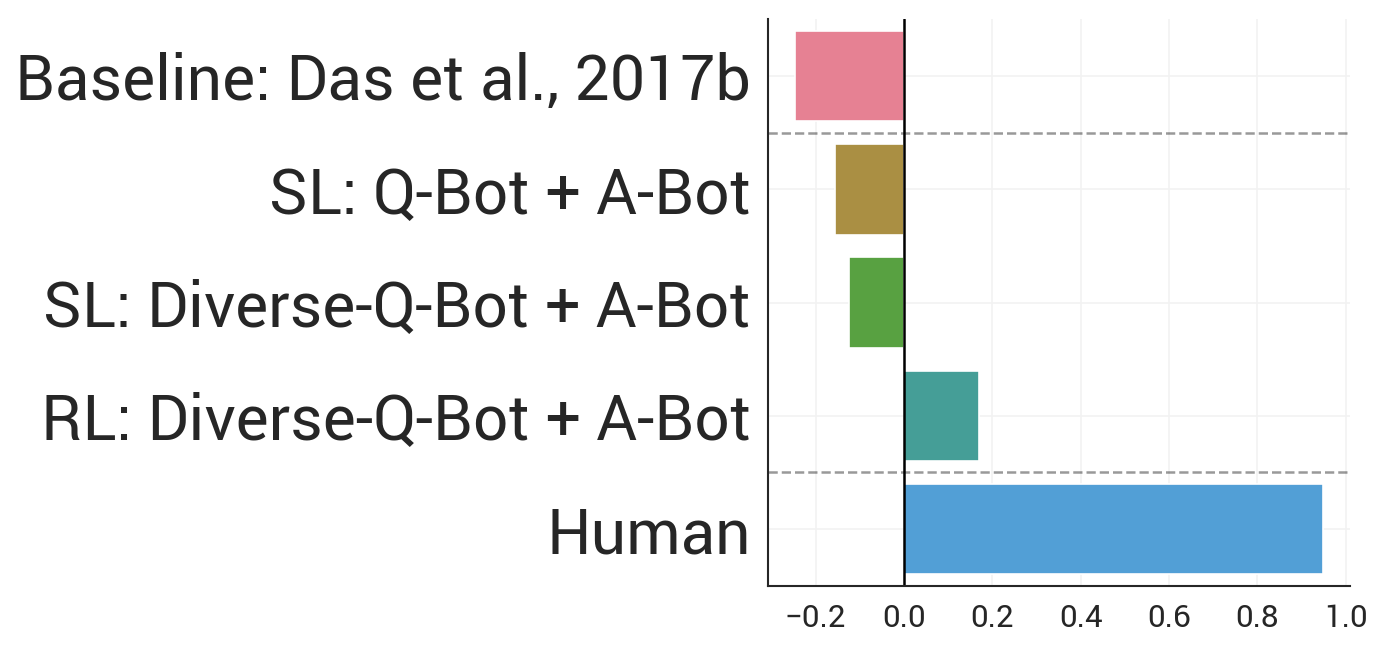}
            \vspace{5pt}
            \caption{{\scriptsize Consistency $\uparrow$}}
            \label{fig:consistency}
        \end{subfigure}
    \end{minipage}\hfill
    \begin{minipage}[b]{0.12\textwidth}
        \begin{subfigure}{\textwidth}
            \centering
            \includegraphics[width=0.9\textwidth]{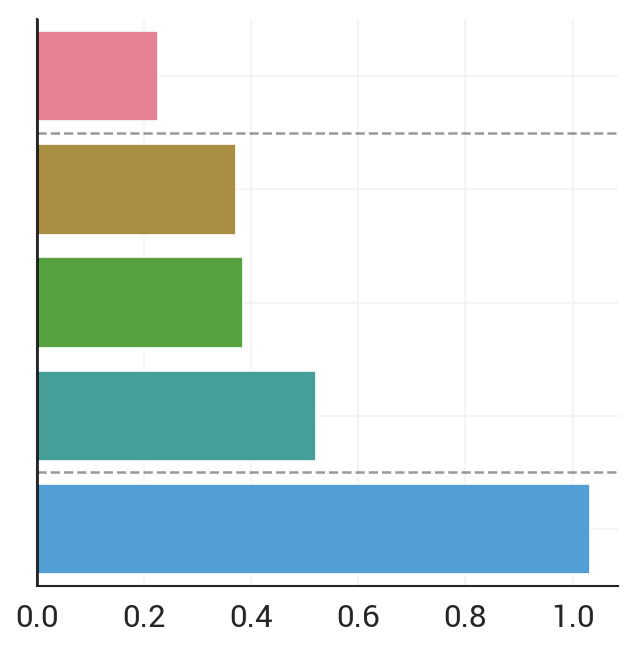}
            \vspace{5pt}
            \caption{{\scriptsize Fluency $\uparrow$}}
            \label{fig:fluency}
        \end{subfigure}
    \end{minipage}\hfill
    \begin{minipage}[b]{0.12\textwidth}
        \begin{subfigure}{\textwidth}
            \centering
            \includegraphics[width=0.9\textwidth]{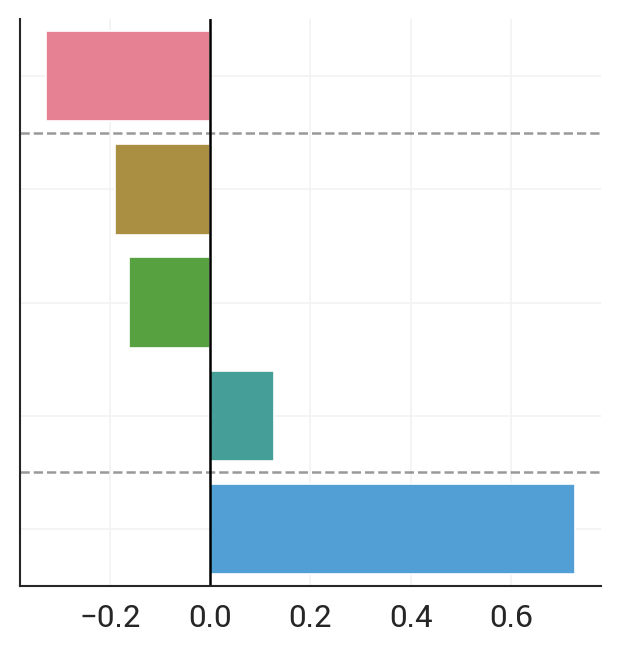}
            \vspace{5pt}
            \caption{{\scriptsize Detail $\uparrow$}}
            \label{fig:detail}
        \end{subfigure}
    \end{minipage}\hfill
    \begin{minipage}[b]{0.12\textwidth}
        \begin{subfigure}{\textwidth}
            \centering
            \includegraphics[width=0.9\textwidth]{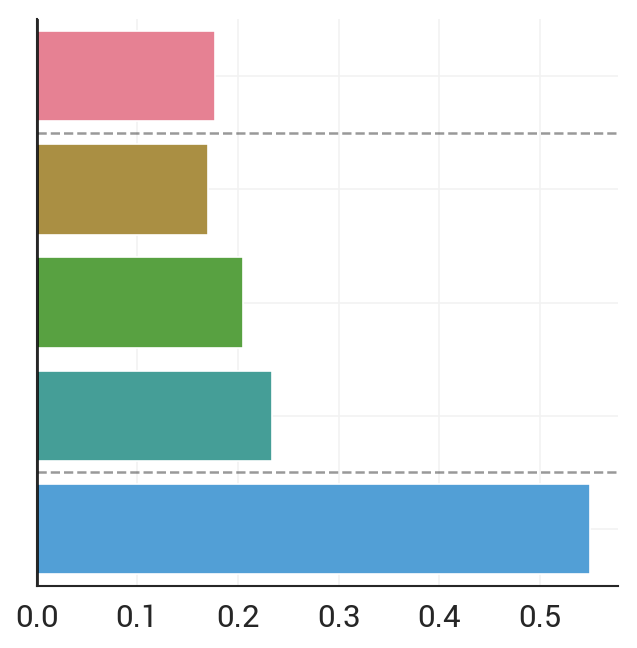}
            \vspace{5pt}
            \caption{{\scriptsize Top-$1$ Acc. $\uparrow$}}
            \label{fig:top-1}
        \end{subfigure}
    \end{minipage}\hfill
    \begin{minipage}[b]{0.33\textwidth}
        \begin{subfigure}{\textwidth}
            \centering
            \includegraphics[width=0.9\textwidth]{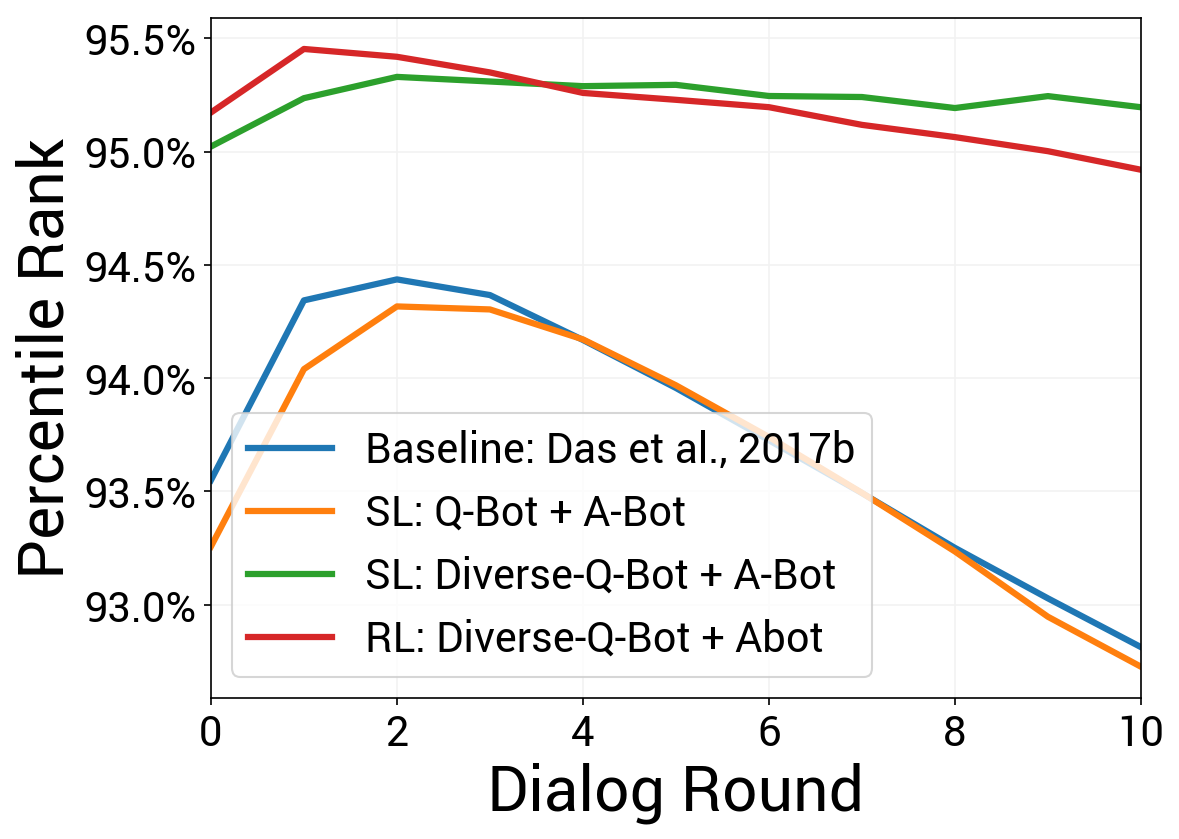}
            \vspace{5pt}
            \caption{{\scriptsize \Qbot-\Abot image-guessing task performance}}
            \label{fig:guesswhich}
        \end{subfigure}
    \end{minipage}\hfill
    \vspace{5pt}
    \caption{(\textbf{a-d}): Human evaluation of \Qbot-\Abot dialog over $50$
        images and $200$ human subjects for each model variant.
        (\textbf{e}): Percentile rank (higher is better) of the true image
            (shown to \Abot) as retrieved using fc7 feature predictions from \Qbot.}
    \label{fig:human_study}
    \vspace{-10pt}
\end{figure*}

While this reinforcement learning approach leads to improved performance on the
image-guessing task than supervised learned agents, it has a few
shortcomings -- 1) image-guessing performance degrades
after a few rounds of dialog (\figref{fig:guesswhich}),
and 2) these improvements over supervised learning do not translate to an improved \Abot,
\ie responses from this Visual Dialog agent are not necessarily better (on
automatic metrics or human judgements), just that the \Qbot-\Abot pair is sufficiently
in sync to do well at image-guessing.

We begin by understanding why this is the case, and find
that \Qbot-\Abot dialog during `self-talk' often tends to be
repetitive \ie the same question-answer pairs get repeated across rounds (\figref{fig:teaser} left).
Since repeated interactions convey no additional information,
image-guessing performance saturates, and even starts to degrade as the
agent forgets useful context from the past.

These repetitions are due to high similarity in \Qbot's context vectors
of successive rounds driving question generation (\figref{fig:similar_embeddings}).
To address this, we devise a smooth-L1 penalty that penalizes similarity
in successive state vectors (\secref{sec:diversity}).
This incentivizes \Qbot to ask diverse questions, thus reducing
repetitions and in turn enabling \Abot to explore a larger part of the state space during
RL~\ie be exposed to more visual concepts to talk about, and varied questions to answer (\secref{sec:results}).

Note that a trivial failure mode with this penalty is for \Qbot to start generating
diverse but totally image-irrelevant questions, which are not useful for the
image-guessing task. A good balance between diversity and image-relevance in \Qbot's
questions is necessary to improve at this task.

We extensively evaluate each component of our approach against prior work and baselines:
\begin{itemize}[leftmargin=*]
\item \Qbot on diversity and image-relevance of generated questions during \Qbot-\Abot self-talk.
We find that diverse-\Qbot asks \emph{more novel questions} while still being image-relevant.
\item \Qbot-\Abot self-talk on consistency, fluency, level of detail, and human-interpretability,
through automatic metrics and human studies.
We find that diverse-\Qbot-\Abot dialog after RL is \emph{more consistent, fluent, and detailed}.
\item \Abot on precision and recall of generated answers on the VisDial dataset~\cite{visdial}
~\ie quality of answers to human questions.
Training diverse-\Qbot $+$ \Abot with RL does not lead to a drop in accuracy on VisDial.
\end{itemize}

\section{Preliminaries}
\label{sec:prelim}

We operate in the same setting as~\citet{visdial_rl}
-- an image-guessing task between a questioner (\Qbot) and an
answerer (\Abot) -- where \Qbot has to guess the image
\Abot has access to by asking questions in multi-round dialog.

We adopt the same training paradigm\footnote{\href{https://github.com/batra-mlp-lab/visdial-rl}{{\tt github.com/batra-mlp-lab/visdial-rl}}}
consisting of 1) a supervised pre-training
stage where \Qbot and \Abot are trained with Maximum Likelihood Estimation
objectives on the VisDial dataset~\cite{visdial}, and 2) a self-talk RL
finetuning stage where \Qbot and \Abot interact with each other and the agents
are rewarded for each successive exchange based on
incremental improvements in guessing the unseen image. We learn parameterized
policies $\pi_{\theta_Q}(q_t|s_{t-1}^Q)$ and $\pi_{\theta_A}(a_t|s_t^A)$
for \Qbot and \Abot respectively which decide what tokens to utter (\textit{actions}:
question $q_t$ and answer $a_t$ at every dialog round $t$) conditioned on the
context available to the agent (\textit{state} representations: $s_{t-1}^Q,s_t^A$).
\Qbot additionally makes an image feature prediction $\hat y_{t}$ at every dialog round (treated as a deterministic \textit{action} with a continuous action space), and the
reward is $r_t = ||y^{gt} - \hat y_{t-1}||_2^2 - ||y^{gt} - \hat y_{t}||_2^2$, \ie
change in distance to the true representation $y^{gt}$ before and after a dialog round.
We use REINFORCE~\cite{williams1992simple}
to update agent parameters,~\ie \Qbot and \Abot are respectively updated with
$\mathbb{E}_{\pi_Q, \pi_A}[r_t \nabla_{\theta_Q} \log \pi_Q(q_t|s_{t-1}^Q)]$ and
$\mathbb{E}_{\pi_Q, \pi_A}[r_t \nabla_{\theta_A} \log \pi_A(a_t|s_t^A)]$ as gradients.

Our transition from supervised to RL is gradual --
we supervise for $N$ rounds and have policy-gradient
updates for the remaining $10-N$, starting from $N=9$ till $N=4$,
one round at a time. After reaching $N=4$, repeating this procedure
from $N=9$ led to further (marginal) improvements.

Both \Qbot and \Abot are modeled as Hierarchical Recurrent Encoder-Decoder architectures~\cite{serban_acl16}.
\Qbot's fc$7$~\cite{simonyan_iclr15} feature prediction of the unseen image ($\hat y$)
is conditioned on the dialog history so far ($s_{t-1}^Q$) via a regression
head\footnote{Please refer to appendix for architecture details.}.

\begin{table*}[t!]
    \resizebox{\textwidth}{!}{
        \begin{tabular}{@{}lccccccccc@{}}
            & \multicolumn{7}{c}{Diversity} && \multicolumn{1}{c}{Relevance} \\[0.01in]
            \cline{2-8} \cline{10-10} \\
            & \# Novel questions $\uparrow$ & \# Unique questions $\uparrow$ & Mutual overlap $\downarrow$
            & Ent-$1$ $\uparrow$ & Ent-$2$ $\uparrow$             & Dist-$1$ $\uparrow$ & Dist-$2$ $\uparrow$
            && Negative log likelihood $\downarrow$ \\[0.05in]
            \toprule
            Baseline: \citet{visdial_rl}       &      $71$ & $6.70$ {\scriptsize $\pm$ $0.07$} & $0.58$ {\scriptsize $\pm$ $0.01$} & $2.72$ {\scriptsize $\pm$ $0.01$} & $3.03$ {\scriptsize $\pm$ $0.02$} &$0.35$ {\scriptsize $\pm$ $0.0$} & $0.43$ {\scriptsize $\pm$ $0.0$}  && $\mathbf{9.94}$ \\
            \cdashline{1-10}

            SL: \Qbot $+$ \Abot                     &  $ 51 $  &  $ 6.57 $  {\scriptsize $\pm$ $0.07$} & $ 0.60 $  {\scriptsize $\pm$  $0.01$}  &  $ 2.70 $
                {\scriptsize $\pm$ $0.01$}  & $3.00$  {\scriptsize $\pm$ $0.02$} & $0.34 $ {\scriptsize $\pm$ $0.0$} & $0.42 $ {\scriptsize $\pm$ $0.0$}  &&  $ 10.05 $ \\
            SL: Diverse-\Qbot $+$ \Abot             & $146$ & $7.45$ {\scriptsize $\pm$ $0.07$} & $0.51$ {\scriptsize $\pm$ $0.01$} & $2.82$ {\scriptsize $\pm$ $0.01$} & $3.18$ {\scriptsize $\pm$ $0.01$} & $0.38$ {\scriptsize $\pm$ $0.0$} & $0.48$ {\scriptsize $\pm$ $0.0$} && $10.10$ \\
            RL: Diverse-\Qbot $+$ \Abot & $\mathbf{449}$ & $\mathbf{8.19}$ {\scriptsize $\pm$ $0.06$} & $\mathbf{0.41}$ {\scriptsize $\pm$ $0.01$} & $\mathbf{2.90}$ {\scriptsize $\pm$ $0.01$} & $\mathbf{3.31}$ {\scriptsize $\pm$ $0.01$} & $\mathbf{0.40}$ {\scriptsize $\pm$ $0.0$} & $\mathbf{0.53}$ {\scriptsize $\pm$ $0.0$} && $10.80$ \\

            \bottomrule
        \end{tabular}
    }
    \vspace{5pt}
    \caption{\Qbot diversity and relevance on v1.0 val.
        $\uparrow$ indicates higher is better.
        $\downarrow$ indicates lower is better.}
    \label{tab:div_rel_res}
    \vspace{-10pt}
\end{table*}

\begin{table*}[ht!]
    \resizebox{\textwidth}{!}{
        \begin{tabular}{@{}lccccccccccccc@{}}
            & \multicolumn{6}{c}{v$1.0$ val} & & \multicolumn{6}{c}{v$1.0$ test-std} \\[0.01in]
            \cline{2-7} \cline{9-14} \\
            & NDCG $\uparrow$ & MRR $\uparrow$ & R@$1$ $\uparrow$ & R@$5$ $\uparrow$ & R@$10$ $\uparrow$ & Mean Rank $\downarrow$
            && NDCG $\uparrow$ & MRR $\uparrow$ & R@$1$ $\uparrow$ & R@$5$ $\uparrow$ & R@$10$ $\uparrow$ & Mean Rank $\downarrow$ \\[0.05in]
            \toprule

            Baseline: \citet{visdial_rl}                            & $53.76$ & $46.35$ & $36.22$ & $56.15$&    $62.41$&    $19.34$ && $51.60$ & $45.67$ & $35.05$ & $56.30$ & $63.25$ & $19.15$ \\
            \cdashline{1-14}

            SL: \Abot                                        & $53.10$ & $46.21$ & $36.11$ & $55.82$ & $62.22$ & $19.58$ && $51.18$ & $45.43$ & $34.88$
            & $55.65$ & $63.20$ & $19.16$\\

            RL: \Abot (finetuned with Diverse-\Qbot)           & $53.91$ & $46.46$  & $36.31$ & $56.26$   & $62.53$ & $19.35$ && $51.67$ & $45.64$ & $34.85$ & $56.55$ & $63.43$ & $18.96$ \\

            \bottomrule
        \end{tabular}
    }
    \vspace{5pt}
    \caption{\Abot performance on VisDial v$1.0$~\cite{visdial}.
        $\uparrow$ indicates higher is better.
        $\downarrow$ indicates lower is better.}
    \label{tab:visdial_res}
    \vspace{-10pt}
\end{table*}

\section{Smooth-L1 Penalty on Question Repetition}
\label{sec:diversity}

Our goal is to encourage \Qbot to ask a diverse set of questions so that
when \Abot is exposed to the same during RL finetuning, it is better able to
explore its state space\footnote{\Abot's state-space is characterized by a representation of the question, image, and the dialog history so far.}.
Furthermore, asking diverse questions allows
\Qbot-\Abot exchanges across rounds to be more informative of the image,
thus more useful for the image-guessing task.

We observe that agents trained using the paradigm proposed by~\citet{visdial_rl}
suffer from repetition of \textit{context} across multiple rounds of dialog --
similar dialog state embeddings across multiple rounds leading to
repeated utterances and similar predicted image representations, which consequently
further increases similarity in state embeddings.~\figref{fig:similar_embeddings}
shows increasing $\cos(s_{t-1}^Q, s_{t}^Q)$ across dialog rounds for~\citet{visdial_rl}.

To encourage \Qbot to ask diverse questions, we propose a simple auxiliary
loss that penalizes similar dialog state embeddings.
Specifically, given \Qbot states -- $s_{t-1}^Q,s_t^Q$ --
in addition to maximizing likelihood of question (during supervised pre-training),
or image-guessing reward (during self-talk RL finetuning), we
maximize a smooth-L1 penalty on
$\Delta_t = \texttt{abs}(\left\lVert s_{t-1}^Q \right\rVert_2 - \left\lVert s_t^Q \right\rVert_2)$,
\begin{equation}
f(\Delta_t) =
\begin{cases}
		0.5 \Delta_t^2 	 & \text{if $\Delta_t  < 0.1$} \\
	    0.1(\Delta_t  - 0.05) & \text{otherwise}
\end{cases}
\label{eqn:diversity}
\end{equation}

resulting in $\sum_{t=2}^{N} f(\Delta_t)$ as an additional term in the overall
objective ($N =$ no. of dialog rounds).

Note that in order to maximize this penalty, \Qbot has to push
$s_{t-1}^Q$ and $s_{t}^Q$ further apart, which can only happen if
$s_{t-1}^Q$ is updated using a question-answer pair that is different from the
previous exchange, thus overall forcing \Qbot to ask different questions in
successive dialog rounds. Similar diversity objectives have also been explored
in~\citet{li_emnlp16} as reward heuristics.

Before arriving at~\eqnref{eqn:diversity}, and following~\figref{fig:similar_embeddings},
we also experimented with directly
minimizing cosine similarity, $\cos(s_{t-1}^Q, s_{t}^Q)$. This led to the network
learning large biases to flip the direction of successive $s_{t-1}^Q$ vectors
(without affecting norms), leading to question repetitions in
alternating rounds.

\section{Experiments}
\label{sec:experiments}

\textbf{Baselines and ablations}. To understand the effect of the proposed penalty,
we compare our full approach -- `{\tt RL: Diverse-\Qbot + \Abot}' -- with
the baseline setup in~\citet{visdial_rl}, as well as several ablations  --
1) `{\tt SL: \Qbot + \Abot}': supervised
agents (\ie trained on VisDial data under MLE, no RL, no smooth-L1 penalty).
Comparing to this quantifies how much our penalty + RL helps.
2) `{\tt  SL: Diverse-\Qbot + \Abot}': supervised agents where \Qbot is trained
with the smooth-L1 penalty. This quantifies gains from RL.

\textbf{Automatic Metrics.}
To evaluate \Qbot's \textit{diversity} (Table~\ref{tab:div_rel_res}), we generate
\Qbot-\Abot dialogs (with beam size = 5) for $10$ rounds on
VisDial v$1.0$ val and compute 1) Novel Questions: the number of new
questions (via string matching) in the generated dialog not seen during training,
2) Unique Questions: no. of unique questions per dialog instance (so $\leq10$),
3) Dist-n and Ent-n~\cite{zhang_nips18,li_naacl16}:
the number and entropy of unique n-grams in the generated questions normalized by
the total number of tokens, and 4) Mutual Overlap~\cite{deshpande2019fast}:
BLEU-4 overlap of every question in the generated $10$-round dialog with the other
$9$ questions, followed by averaging these $10$ numbers. To measure \Qbot's \textit{relevance},
we report the negative log-likelihood under the model of human questions from VisDial.
We evaluate \Abot's answers to human questions from the VisDial dataset
on the retrieval metrics introduced by~\citet{visdial} (Table~\ref{tab:visdial_res}).
Finally, we also evaluate performance of the \Qbot-\Abot pair at image-guessing (\figref{fig:guesswhich}),
which is the downstream task they are trained with RL for.

\textbf{Human Studies}.
To evaluate how human-understandable \Qbot-\Abot
dialogs are, we conducted a study where we showed humans these dialogs
(from our agents as well as baselines), along with a pool of $16$ images from
the VisDial v$1.0$ test-std split -- consisting of the unseen image, $5$ nearest neighbors
(in fc$7$ space), and $10$ random images -- and asked humans to pick their top-5 guesses for the
unseen image. Our hypothesis was that if questions are more diverse, the dialog
will be more image-informative, and so humans should be able to better guess which
image was being talked about. We report top-$1$ accuracy of true image
in human guesses. We also asked humans to rate \Qbot-\Abot dialog on consistency,
fluency, and level of detail on a $5$-point Likert scale.

\section{Implementation Details}
We used beam search with a beam size of 5 during self-talk between all \Qbot-\Abot variants.
NDCG scores on the v1.0 val split and the total SL loss (on the same split)
were used to select the best SL \Abot and \Qbot checkpoints respectively.
We used a dropout rate of $0.5$ for all SL-pretraining experiments and
no dropout for RL-finetuning. We used Adam~\cite{kingma_iclr15} with a learning
rate of $10^{-3}$ decayed by ${\sim}0.25$ every epoch, upto a minimum of $5\times10^{-5}$.

The objective for training Diverse-\Qbot was a sum of the smooth-L1 penalty
(introduced in \secref{sec:diversity}), cross entropy loss, and L2 loss between
the regression head output and the fc7~\cite{simonyan_iclr15} embedding of the image.
We observed that coefficients in the range of $1e^{-3}$ to $1e^{-5}$ worked best
for the smooth-L1 penalty. We also observed that training for a large
number of epochs ($\sim$80) with the above mentioned range of coefficient values
led to the best results.

\section{Results}
\label{sec:results}

\begin{itemize}[leftmargin=*]
	\item \textbf{\Qbot's diversity} (Table~\ref{tab:div_rel_res}):
    The question-repetition penalty consistently increases diversity (in both SL
    and RL) over the baseline!
    {\tt RL: Diverse-\Qbot}
    asks \char`\~$1.5$ more unique questions on average than~\citet{visdial_rl} ($6.70\rightarrow8.19$)
    for every $10$-round dialog, \char`\~6.3x more novel questions ($71\rightarrow449$),
    and a higher fraction and entropy of unique generated n-grams, while still
    staying comparably relevant (NLL).

    \item \textbf{\Abot on VisDial} (Table.~\ref{tab:visdial_res}):
    {\tt RL: \Abot} outperforms {\tt SL: \Abot}, but does not statistically improve
    over the baseline on answering human questions from
    VisDial\footnote{This is consistent with trends in~\citet{visdial_rl}.}
    (on v$1.0$ val \& test-std).

    \item \textbf{Image-guessing task} (\figref{fig:guesswhich}):
    {\tt Diverse-\Qbot + \Abot} (SL and RL) significantly outperform
    the baseline on percentile rank of ground-truth image as retrieved
    using \Qbot's fc$7$ prediction. Thus, the question-repetition
    penalty leads to a more informative communication protocol.

	\item \textbf{Human studies} (\figref{fig:human_study}):
    Humans judged {\tt RL: Diverse-\Qbot + \Abot} dialog significantly
    more consistent (fewer conflicting exchanges), fluent (fewer grammatical
    errors), and detailed (more image-informative) over the baseline
    and supervised learning. This is an important result.
    Performance on GuessWhich, together with these dialog quality judgements from humans
    show that agents trained with our approach develop a more effective communication protocol
    for the downstream image-guessing task, while still not deviating off English, which is
    a common pitfall when training dialog agents with RL~\cite{kottur_emnlp17,lewis_emnlp17}.

\end{itemize}

Note that since our penalty (Eqn.~\ref{eqn:diversity})
is structured to avoid repetition across successive
rounds, one possible failure mode is that \Qbot learns
to ask the same question every alternate dialog round
(at $t$ and $t+2$). Empirically, we find that this
happens ${\sim}15\%$ of times
($2490$ times out of ${\sim}16.5$k question pairs)
on v1.0 val for RL: Diverse \Qbot + \Abot compared to ${\sim}22\%$ for SL: \Qbot + \Abot.
This observation, combined with the fact that
Diverse-\Qbot asks $\sim$1.6 more unique questions relative to SL: \Qbot across 10 rounds
suggests that simply incentivizing diversity in successive rounds works well empirically.
We hypothesize that this is because repeating questions every other round or other such
strategies to game our repetition penalty is fairly specific behavior that is likely hard
for models to learn given the large space of questions \Qbot could potentially ask.

\section{Related Work}
\label{sec:related}

Our work is related to prior work in visual dialog~\cite{visdial,vries_cvpr17}
and modeling diversity in text-only dialog~\cite{zhang_nips18,li_naacl16,li_emnlp16}.

Closest to our setting is work on using conditional variational autoencoders for
self-talk in visual dialog~\cite{massiceti_cvpr18}, where diversity is not explicitly modeled but
is measured via metrics specific to the proposed architecture.

Adding constraints to generate a diverse set of natural language dialog
responses have previously been explored in~\citet{zhang_nips18} via adversarial information maximization,
in~\citet{gao_arxiv19} by jointly modeling diversity and relevance in a shared latent space,
and in~\citet{li_naacl16} using a maximum mutual information criterion.
In contrast, we are interested in diversity at the level of the entire dialog
(instead of a single round) -- reducing repetitions in QA pairs across multiple rounds.
Our
repetition
penalty is partly inspired by the `Information Flow' constraint in~\citet{li_emnlp16}.
As detailed in~\secref{sec:prelim}, we experimented with similar forms of the penalty and eventually settled on smooth-L1.
To the best of our knowledge, we are the first to explicitly model diversity as a constraint in visual dialog.
\section{Conclusions \& Future Work}

We devised an auxiliary objective for training
generative Visual Dialog agents with RL, that
incentivizes \Qbot to ask diverse questions.
This reduces repetitions in \Qbot-\Abot dialog during RL self-talk,
and in turn enables \Abot to be exposed to a larger state space.
Through extensive evaluations,
we demonstrate that our \Qbot-\Abot pair
has significantly more diverse dialog
while still being image-relevant, better downstream task performance,
and higher consistency, fluency, and level of detail than
baselines. Our code is publicly available: \href{https://github.com/vmurahari3/visdial-diversity}{{\tt github.com/vmurahari3/visdial-diversity}}.
We hope this serves as a robust base for furthering progress on training visual dialog
agents with RL for other grounded language games, adapting to
learn to talk about novel visual domains, \etc.

\section{Acknowledgements}
\label{sec:ack}

We thank Nirbhay Modhe and Viraj Prabhu for the PyTorch implementation~\cite{modhe_github18} of~\citet{visdial_rl} that we built on, and Jiasen Lu for helpful discussions. The Georgia Tech effort is supported in part by NSF, AFRL, DARPA, ONR YIPs, ARO PECASE. AD is supported in part by fellowships from Facebook, Adobe, and Snap Inc.
The views and conclusions contained herein are those of the authors
and should not be interpreted as necessarily representing the official policies
or endorsements, either expressed or implied, of the US Government, or any
sponsor.

% We thank Nirbhay Modhe and Viraj Prabhu for the PyTorch implementation~\cite{modhe_github18}
% of~\citet{visdial_rl} that we built on, and Jiasen Lu for helpful discussions.
% The Georgia Tech effort is supported in part by NSF, AFRL, DARPA, ONR YIPs,
% ARO PECASE. AD is supported in part by fellowships from Facebook, Adobe, and Snap Inc.
% The views and conclusions contained herein are those of the authors
% and should not be interpreted as necessarily representing the official policies
% or endorsements, either expressed or implied, of the US Government, or any
% sponsor.
\bibliography{strings,main}
\bibliographystyle{acl_natbib}
\newpage
\section{Appendix}

\subsection{Qualitative Output}
The dialog generated by self-talk between different SL model variants is provided in Table.~\ref{table:sl} and different RL model variants is provided in Table.~\ref{table:rl}. We observe the variants with Diverse-\Qbot tend to generate more diverse, image relevant and fluent dialog.

\subsection{Experiments with Diverse-\Abot}
\begin{figure}
\centering
        \includegraphics[width=\columnwidth]{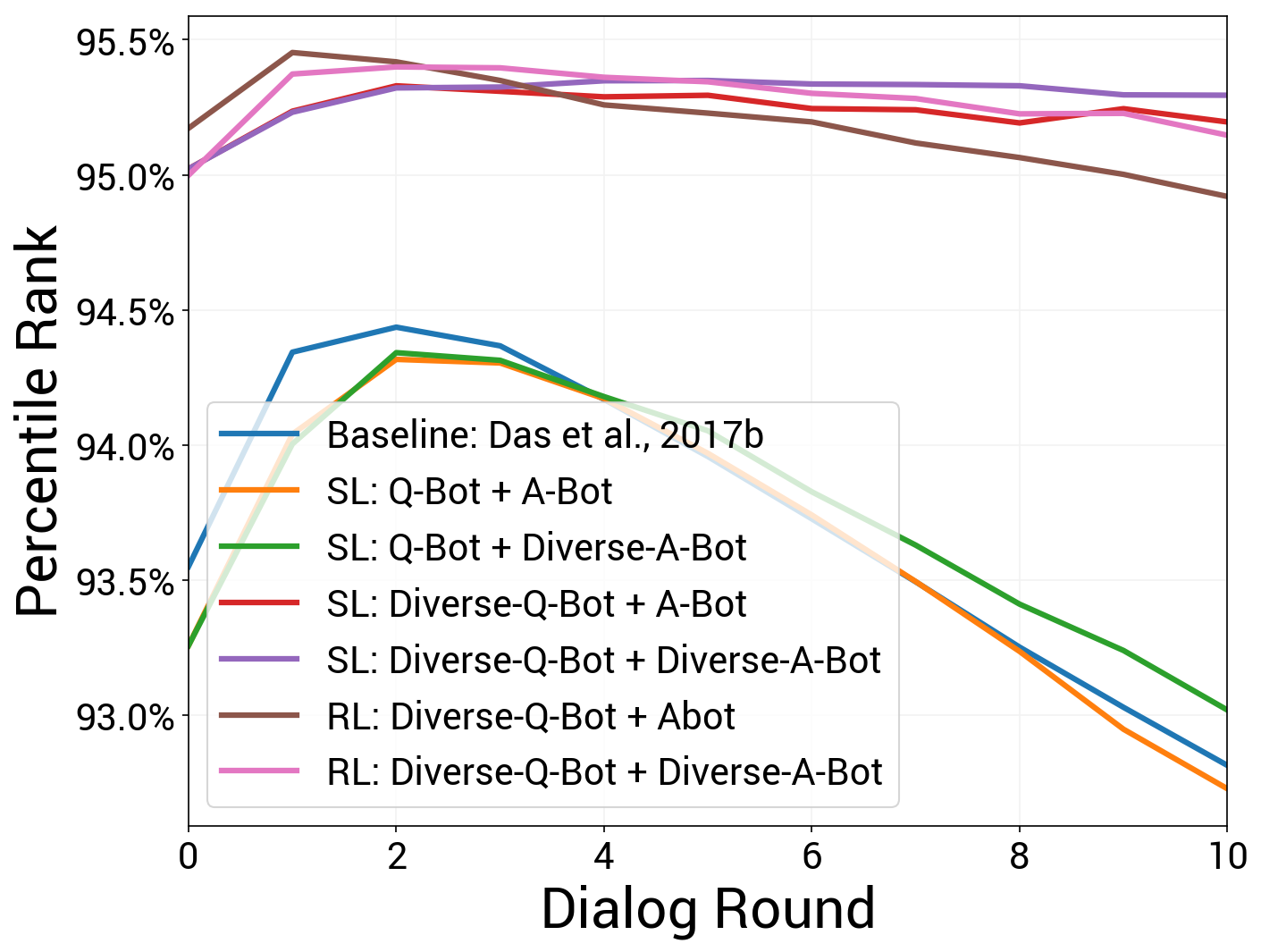}
    \caption{Performance on the image-guessing task. Percentile rank (higher is better) of the true image (shown to A-BOT) as retrieved using fc7 image feature predictions from Q-BOT.}
    \label{fig:guesswhich_supp}
\end{figure}

We also ran experiments where we used the
repetition penalty (Eqn. (1) in the main paper)
during SL pre-training of \Abot.
In Tables~\ref{tab:visdial_res_supp} and~\ref{tab:div_rel_res_supp} we report results on the \Abot retrieval metrics~\cite{visdial} and the diversity metrics respectively.
In Fig.~\ref{fig:guesswhich_supp}, we provide the performance of various \Qbot variants paired with this Diverse-\Abot on the image-guessing task.

We note that retrieval metrics for Diverse-\Abot are better than vanilla SL: \Abot. Therefore, this repetition penalty does help significantly during
supervised
pre-training. However, finetuning
the Diverse-\Abot via RL does not lead to significant improvements.

We note that for the diversity metrics on \Qbot-\Abot dialog, self-talk with Diverse-\Abot does not change the diversity metrics significantly. We observe the same trend in image-guessing performance as well.\\

\subsection{Model Architecture details}

\begin{figure*}
        \includegraphics[scale=0.3]{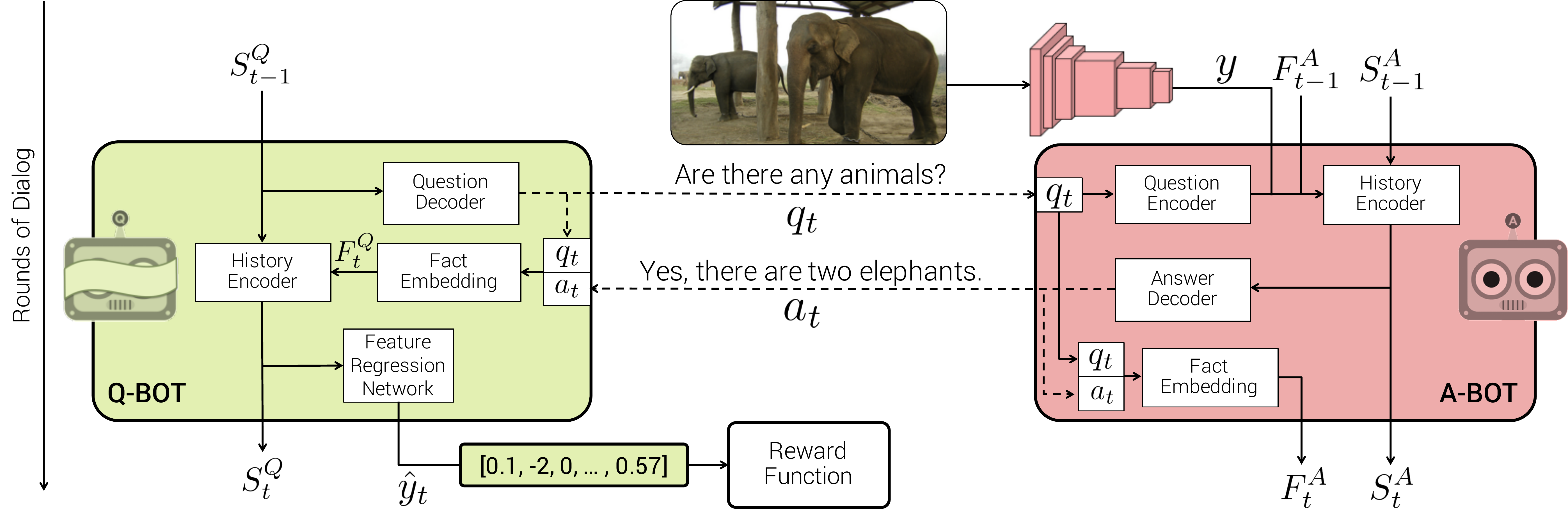}
    \caption{Model architecture diagram borrowed from \cite{visdial_rl} with permission.}
    \label{fig:hre}
\end{figure*}

    We use a Hierarchical Recurrent Encoder (HRE) for representing dialog context. In this encoder the image representation is concatenated with every question word when fed to the LSTM. We then encode each QA-pair in the dialog history with another LSTM with shared weights.
    The image-question representation, computed for every round from $1$ through $t$, is concatenated with history representation from the previous round. This gives us a set of $t$ question-history vectors for t rounds. These vectors are fed as input to a dialog-level LSTM, whose output state at $t$ is used to decode the response to $Q_t$. \ref{fig:hre} shows the model architecture of the HRE. \\

All LSTMs are 2-layered with $512$-dim hidden states.
 We learn $300$-dim embeddings for words and images.
 These word embeddings are shared across question, history, and decoder LSTMs.

\begin{table*}[ht!]
	\resizebox{\textwidth}{!}{
		\begin{tabular}{@{}lccccccccccccc@{}}
			& \multicolumn{6}{c}{v$1.0$ val} & & \multicolumn{6}{c}{v$1.0$ test-std} \\[0.01in]
			\cline{2-7} \cline{9-14} \\
			& NDCG $\uparrow$ & MRR $\uparrow$ & R@$1$ $\uparrow$ & R@$5$ $\uparrow$ & R@$10$ $\uparrow$ & Mean Rank $\downarrow$
			&& NDCG $\uparrow$ & MRR $\uparrow$ & R@$1$ $\uparrow$ & R@$5$ $\uparrow$ & R@$10$ $\uparrow$ & Mean Rank $\downarrow$ \\[0.05in]
			\toprule

			Baseline: \citet{visdial_rl}                            & $53.76$ & $46.35$ & $36.22$ & $56.15$&    $62.41$&    $19.34$ && $51.60$ & $45.67$ & $35.05$ & $56.30$ & $63.25$ & $19.15$ \\

			\cdashline{1-14}

			Diverse SL \Abot                                        & $53.82$ & $46.55$ & $36.46$ & $56.16$ &   $62.68$ &   $19.34$
			&& $51.77$ & $45.98$ & $35.58$ & $56.33$ & $63.08$ & $18.97$\\

			Diverse RL \Abot (Finetuned with Diverse-\Abot)              & $53.94$ & $46.67$ & $36.54$ &    $56.34$ &   $62.99$ &   $19.24$
			&& $51.80$ & $45.70$ & $34.85$ & $56.33$ & $63.90$ & $18.88$ \\

			\bottomrule
		\end{tabular}
	}
	\vspace{5pt}
	\caption{\Abot performance on the VisDial v$1.0$ dataset~\cite{visdial}.
		Arrow indicates direction of better performance.}
	\label{tab:visdial_res_supp}
\end{table*}

\begin{table*}[t!]
	\resizebox{\textwidth}{!}{
		\begin{tabular}{@{}lccccccccc@{}}
			& \multicolumn{7}{c}{Diversity} && \multicolumn{1}{c}{Relevance} \\[0.01in]
			\cline{2-8} \cline{10-10} \\
			& \# Novel questions $\uparrow$ & \# Unique questions $\uparrow$ & Mutual overlap $\downarrow$
			& Ent-$1$ $\uparrow$ & Ent-$2$ $\uparrow$             & Dist-$1$ $\uparrow$ & Dist-$2$ $\uparrow$
			&& Negative log likelihood $\downarrow$  \\[0.05in]
			\toprule
			Baseline: \citet{visdial_rl}       &      $71$ & $6.70$ $\pm$ $0.07$ & $0.58$ $\pm$ $0.01$ & $2.72$ $\pm$ $0.01$ & $3.03$ $\pm$ $0.02$ &$0.35$ $\pm$ $0.0$ & $0.43$ $\pm$ $0.0$  && $9.94$  \\
			\cdashline{1-10}

			SL: \Qbot $+$ Diverse-\Abot             & $53$ & $6.58$ $\pm$ $0.07$ & $0.60$ $\pm$ $0.01$ & $2.71$ $\pm$ $0.01$ & $3.00$ $\pm$ $0.01$ & $0.34$ $\pm$ $0.0$ & $0.42$ $\pm$ $0.0$ && $10.05$ \\
			SL: Diverse-\Qbot $+$ Diverse-\Abot & $136$ & $7.42$ $\pm$ $0.07$ & $0.50$ $\pm$ $0.01$ & $2.81$ $\pm$ $0.01$ & $3.18$ $\pm$ $0.01$ & $0.38$ $\pm$ $0.0$ & $0.48$ $\pm$ $0.0$ && $10.10$   \\

			RL: Diverse-\Qbot $+$ Diverse-\Abot & $288$ & $7.87$ $\pm$ $0.06$ & $0.46$ $\pm$ $0.01$ & $2.86$ $\pm$ $0.01$ & $3.24$ $\pm$ $0.01$ & $0.39$ $\pm$ $0.0$ & $0.51$ $\pm$ $0.0$ && $10.32$  \\

			\bottomrule
		\end{tabular}
	}
	\vspace{5pt}
	\caption{\Qbot diversity and relevance on v1.0 val.
		$\uparrow$ indicates higher is better.
		$\downarrow$ indicates lower is better.}
	\label{tab:div_rel_res_supp}
\end{table*}

\begin{table*}[p]
    \vspace{-5pt}
	\renewcommand*{\arraystretch}{0.8}
    \setlength{\tabcolsep}{8px}
    \begin{center}\hspace{-5.5pt}
    \resizebox{2.10\columnwidth}{!}{
    \begin{tabular}{ p{2.85cm}  p{4cm}  p{5.3cm}  p{6.5cm}  }
    \toprule
    {\small Image + Caption} & {\small GT dialog~\cite{visdial}} & {\small \Qbot-\Abot dialog} & {\small Diverse-\Qbot-\Abot dialog} \\
    \cmidrule(r){1-1}\cmidrule(lr){2-2}\cmidrule(l){3-3}\cmidrule(l){4-4}
    \raisebox{0pt}{\includegraphics[width=2.7cm]{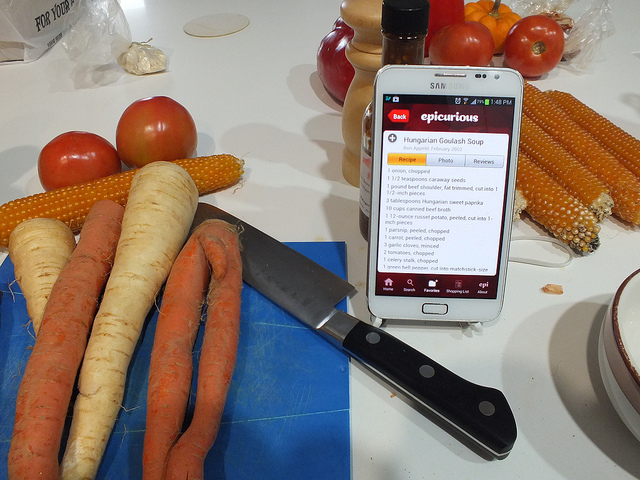}}\newline\vspace{-12pt}
    \begin{minipage}{10cm}{\scriptsize a cutting board with carrots next to a cell phone.}\end{minipage}
    &
    \resizebox{4cm}{!}{
    \begin{minipage}{1.71in}
    \vspace{-57pt}
    {\tiny
    \begin{compactenum}[\hspace{-5pt}]
    \item what color is the cutting board?   \hfill blue
    \item how big is it  \hfill about the size of a regular laptop screen
    \item how many carrots are there ? \hfill 4
    \item are they the orange kind ?  \hfill 2 are orange , 2 are white
    \item what kind of cell phone is it ? \hfill a samsung
    \item can you see any text on the screen ? \hfill a recipe for UNK UNK soup
    \item do you see the room ?  \hfill just the counter
    \item are there people ? \hfill not visible
    \item what color is the counter ?  \hfill white
    \end{compactenum}}
    \end{minipage}}
    &
    \resizebox{5.3cm}{!}{
    \begin{minipage}{2in}
    \vspace{-44pt}
    {\tiny
    \begin{compactenum}[\hspace{-5pt}]
    \item is the photo in color ? \hfill no
    \item how many people do you see ?  \hfill no people
    \item is this a close up ?  \hfill no
    \item is this a close up ?  \hfill zebra is
    \item is this a close up ? \hfill no
    \item is this a close up ?  \hfill no
    \item is this a close up ?  \hfill no
    \item is this a close up ?\hfill no
    \item is this a close up ? \hfill no
    \item is this a close up ?\hfill no
    \end{compactenum}}
    \end{minipage}}
    &
    \resizebox{6.5cm}{!}{
    \begin{minipage}{2.5in}
    \vspace{-45pt}
    {\tiny
    \begin{compactenum}[\hspace{-5pt}]
\item is the photo in color ? \hfill  1 picture is in color

\item is the photo close up ? \hfill  black and white

\item is the photo in color ? \hfill  no it 's black and white

\item is it a close up ? \hfill  no it 's a close up of the cake

\item can you tell what brand it is ? \hfill  no i ca n't see the brand

\item can you tell if it 's a male or female ? \hfill  no i ca n't tell

\item can you tell if it 's a male or female ? \hfill  no i ca n't tell

\item is this in a kitchen ? \hfill  nope it 's in a home

\item are there any utensils ? \hfill  no

\item are there any utensils ? \hfill  nope
    \end{compactenum}}
    \end{minipage}}
    \\[-2pt]  \midrule\iftrue
    \raisebox{0pt}{\includegraphics[width=2.7cm]{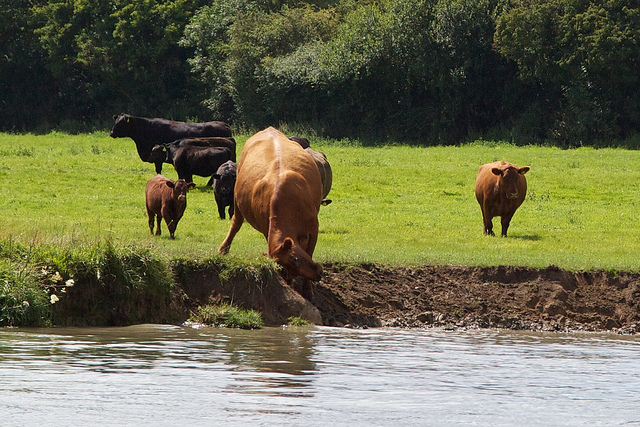}}\newline\vspace{-12pt}
    \begin{minipage}{10cm}{\scriptsize many brown and black cows standing in the grass}\end{minipage}
    &
    \resizebox{4cm}{!}{
    \begin{minipage}{1.6in}
    \vspace{-44pt}
    {\tiny
    \begin{compactenum}[\hspace{-5pt}]
\item is the field large ? \hfill  yes

\item is the grass green or brown ? \hfill  green

\item can you see a barn ? \hfill  no

\item can you see a fence ? \hfill  no

\item can you see a water supply ? \hfill  yes

\item can you see any people ? \hfill  no

\item can you see trees ? \hfill  yes

\item can you see mountains in the background ? \hfill  no

\item is the grass tall or short ? \hfill  short
\end{compactenum}}
    \end{minipage}}
    &
    \resizebox{5.3cm}{!}{
    \begin{minipage}{2in}
    \vspace{-33pt}
    {\tiny
    \begin{compactenum}[\hspace{-5pt}]
\item what color are the cows ? \hfill  they are brown and white

\item how many cows are there ? \hfill  4

\item what color are the cows ? \hfill  they are brown and white

\item can you see the sky ? \hfill  it is not visible

\item do you see any trees ? \hfill  i do not see any trees

\item can you see the sky ? \hfill  green sky

\item are there any clouds ? \hfill  the sky is not visible

\item how many cows are there ? \hfill  2

\item what color are the cows ? \hfill  the cows are brown and white

\item what color are the cows ? \hfill  they are brown    \end{compactenum}}
    \end{minipage}}
    &
    \resizebox{6.5cm}{!}{
    \begin{minipage}{2.6in}
    \vspace{-38pt}
    {\tiny
    \begin{compactenum}[\hspace{-5pt}]
\item do you see any people ? \hfill  1

\item do you see a fence ? \hfill  a small 1

\item how many cows are there ? \hfill  5

\item what color are the cows ? \hfill  black and white

\item can you see the sky ? \hfill  some

\item can you see the sky ? \hfill  nope

\item are there any other animals ? \hfill  not that i can see

\item can you see the sky ? \hfill  not at all

\item are there any trees ? \hfill  in the background

\item can you see the sky ? \hfill  not at all
    \end{compactenum}}
    \end{minipage}}
    \\[-2pt] \midrule
    \fi
    \raisebox{0pt}{\includegraphics[width=2.7cm]{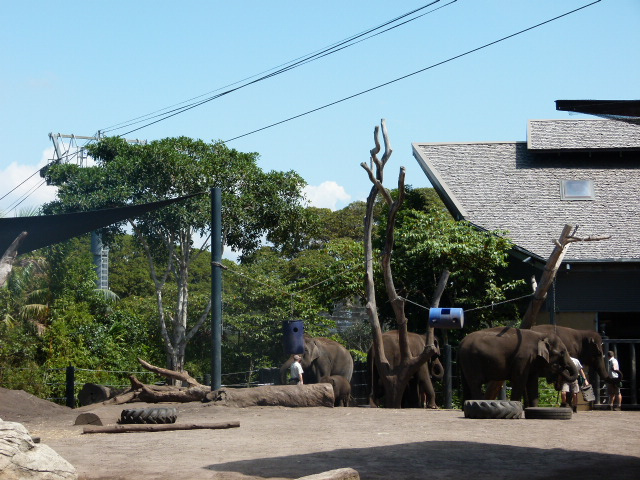}}\newline\vspace{-12pt}
    \begin{minipage}{10cm}{\scriptsize  many elephants standing in an enclosure near a building}\end{minipage}
    &
    \resizebox{4cm}{!}{
    \begin{minipage}{1.5in}
    \vspace{-50pt}
    {\tiny
    \begin{compactenum}[\hspace{-5pt}]
\item what type of building ? \hfill  may be part of a zoo

\item how many elephants ? \hfill  about 4

\item is it sunny ? \hfill  clear day , yes

\item any trees ? \hfill  yes

\item what color is the building ? \hfill  dark grey

\item are there other animals ? \hfill  no

\item are there any people ? \hfill  yes

\item what are they wearing ? \hfill  shorts and shirts

\item how many ? \hfill  3
    \end{compactenum}}
    \end{minipage}}
    &
    \resizebox{5.3cm}{!}{
    \begin{minipage}{2in}
    \vspace{-45pt}
    {\tiny
    \begin{compactenum}[\hspace{-5pt}]
\item how many elephants are there ? \hfill  ca n't tell

\item what color are the elephants ? \hfill  ca n't tell

\item how many elephants are there ? \hfill  ca n't tell

\item what color is the fence ? \hfill  i ca n't tell

\item can you see the sky ? \hfill  i ca n't see the sky

\item are there any trees ? \hfill  just a few

\item does this look like a zoo ? \hfill  1 of them is

\item are there any people ? \hfill  nothing in the picture

\item how many elephants are there ? \hfill  just 1

\item can you see the sky ? \hfill  just a little
    \end{compactenum}}
    \end{minipage}}
    &
    \resizebox{6.5cm}{!}{
    \begin{minipage}{2.5in}
    \vspace{-47pt}
    {\tiny
    \begin{compactenum}[\hspace{-5pt}]
\item how many elephants are there ? \hfill  1

\item any people ? \hfill  ca n't see any

\item male or female ? \hfill  not sure

\item what is the elephant doing ? \hfill  just standing there

\item male or female ? \hfill  there are 2 males and 1 female

\item can you see the sky ? \hfill  1 of it

\item what color is the building ? \hfill  just a black and white picture

\item do you see any trees ? \hfill  there are trees in the background

\item are there any other animals ? \hfill  red and white

\item do you see any trees ? \hfill  i see some in the background
    \end{compactenum}}
    \end{minipage}}
    \\[-2pt] \midrule
    \raisebox{0pt}{\includegraphics[width=2.7cm]{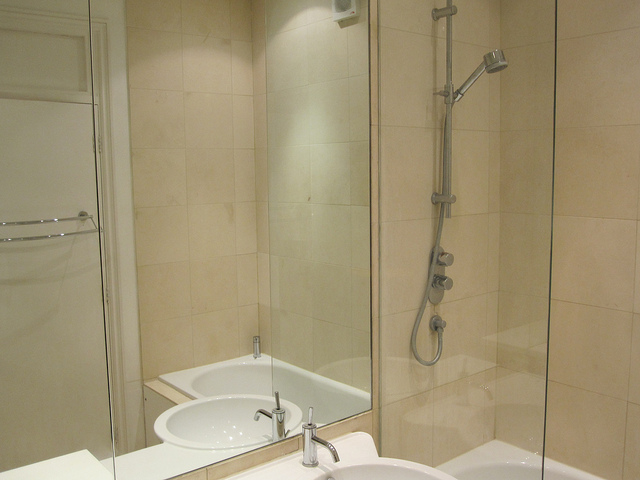}}\newline\vspace{-12pt}
    \begin{minipage}{10cm}{\scriptsize a reflection of the tub is in the bathroom mirror along with the sink , you can see all the room}\end{minipage}
    &
    \resizebox{4cm}{!}{
    \begin{minipage}{1.5in}
    \vspace{-50pt}
    {\tiny
    \begin{compactenum}[\hspace{-5pt}]
\item is the photo in color ? \hfill  yes

\item is there humans in the photo ? \hfill  no

\item is there any animals in the photo ? \hfill  no

\item what color are the walls ? \hfill  tan

\item what color is the sink ? \hfill  white

\item is there any towels ? \hfill  no

\item is there any pictures on the wall ? \hfill  no

\item what color is the tub ? \hfill  white

\item can you see the floor ? \hfill  no
    \end{compactenum}}
    \end{minipage}}
    &
    \resizebox{5.3cm}{!}{
    \begin{minipage}{2.1in}
    \vspace{-50pt}
    {\tiny
    \begin{compactenum}[\hspace{-5pt}]
\item is the photo in color ? \hfill  5 of them are

\item are there any people ? \hfill  it 's a close up so i ca n't see any people

\item what color is the sink ? \hfill  about the same color as the toilet

\item do you see a mirror ? \hfill  no

\item what color is the sink ? \hfill  yes it is white

\item what color is the sink ? \hfill  yes it is white

\item what shape is the mirror ? \hfill  yes , it 's a rectangle

\item are there any towels ? \hfill  he 's on a towel rack

\item are there any towels ? \hfill  i do n't see any towels

\item are there any towels ? \hfill  i do n't see any towels
    \end{compactenum}}
    \end{minipage}}
    &
    \resizebox{6.5cm}{!}{
    \begin{minipage}{2.5in}
    \vspace{-47pt}
    {\tiny
    \begin{compactenum}[\hspace{-5pt}]
\item is the photo in color ? \hfill  no

\item is there any people ? \hfill  5

\item what color walls ? \hfill  i ca n't see the walls

\item is there a mirror ? \hfill  yes

\item is there a mirror ? \hfill  yes

\item is there a mirror ? \hfill  yes

\item are the walls seen ? \hfill  no

\item what color are the walls ? \hfill  no walls

\item what is the floor made of ? \hfill  yes

\item what color is the floor ? \hfill  i ca n't see the floor
    \end{compactenum}}
    \end{minipage}}
    \\[-2pt] \midrule
    \raisebox{0pt}{\includegraphics[width=2.7cm]{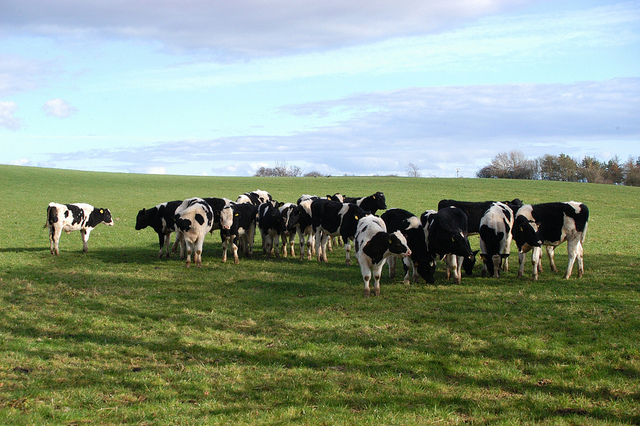}}\newline\vspace{-12pt}
    \begin{minipage}{10cm}{\scriptsize a bunch of cows that are in a grassy field}\end{minipage}
    &
    \resizebox{4cm}{!}{
    \begin{minipage}{1.5in}
    \vspace{-38pt}
    {\tiny
    \begin{compactenum}[\hspace{-5pt}]
\item about how many cows ? \hfill  over a dozen
\item are there any calves ? \hfill  no
\item are they eating ? \hfill  yes
\item is it sunny ? \hfill  no
\item is there a fence around them ? \hfill  no
\item do you see any buildings ? \hfill  no
\item mountains in the background ? \hfill  no
\item is there snow on the ground ? \hfill  no
\item any trees ? \hfill  yes
    \end{compactenum}}
    \end{minipage}}
    &
    \resizebox{5.3cm}{!}{
    \begin{minipage}{2in}
    \vspace{-33pt}
    {\tiny
    \begin{compactenum}[\hspace{-5pt}]
\item is the photo in color ? \hfill  yes
\item is there any people ? \hfill  no
\item is there a fence ? \hfill  no
\item is there a fence ? \hfill  no
\item is there a fence ? \hfill  no
\item is there a fence ? \hfill  no
\item is there a fence ? \hfill  no
\item is there a fence ? \hfill  no
\item is there a fence ? \hfill  no
\item is there a fence ? \hfill  no
    \end{compactenum}}
    \end{minipage}}
    &
    \resizebox{6.5cm}{!}{
    \begin{minipage}{2.5in}
    \vspace{-34pt}
    {\tiny
    \begin{compactenum}[\hspace{-5pt}]
\item how many cows are there ? \hfill  about 10

\item what color are the cows ? \hfill  black and white

\item how many cows are there ? \hfill  about 10

\item what color are the cows ? \hfill  black and white

\item what is the weather like ? \hfill  sunny

\item are there any trees ? \hfill  no

\item are there any people ? \hfill  no

\item are there any trees ? \hfill  no

\item are there any buildings ? \hfill  no

\item are there any trees ? \hfill  yes
    \end{compactenum}}
    \end{minipage}}
    \\[-2pt] \midrule
    \raisebox{0pt}{\includegraphics[width=2cm,height=2cm]{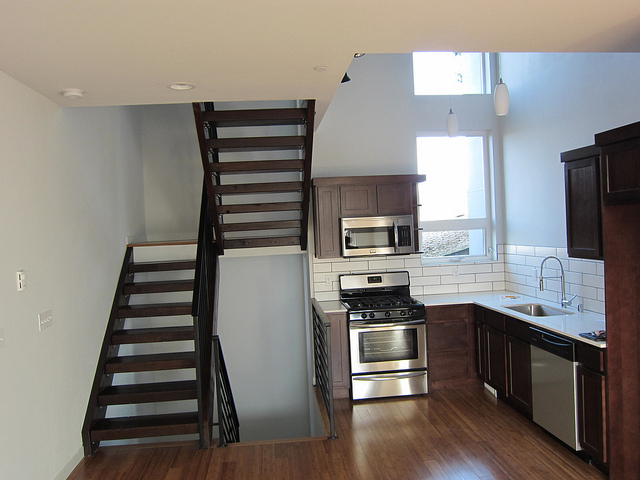}}\newline\vspace{-12pt}
    \begin{minipage}{10cm}{\scriptsize the kitchen is UNK with dark wood cabinets and a nearby staircase}\end{minipage}
    &
    \resizebox{4cm}{!}{
    \begin{minipage}{1.53in}
    \vspace{-45pt}
    {\tiny
    \begin{compactenum}[\hspace{-5pt}]
\item what color are the kitchen walls ? \hfill  white

\item is there a window present ? \hfill  yes

\item are there pictures paintings hanging up ? \hfill  no

\item is the stove visible ? \hfill  yes

\item what color is the stove ? \hfill  black and silver

\item are there any people visible ? \hfill  no

\item what color is the nearby staircase ? \hfill  dark brown

\item is the staircase carpeted ? \hfill  no

\item are the handrails of the staircase visible ? \hfill  yes
    \end{compactenum}}
    \end{minipage}}
    &
    \resizebox{5.3cm}{!}{
    \begin{minipage}{2in}
    \vspace{-43pt}
    {\tiny
    \begin{compactenum}[\hspace{-5pt}]
\item is the photo in color ? \hfill  yes

\item is there any people ? \hfill  no

\item is there a stove ? \hfill  no

\item are there any windows ? \hfill  no

\item can you see the floor ? \hfill  no

\item can you see the floor ? \hfill  no

\item can you see the floor ? \hfill  yes

\item can you see the floor ? \hfill  yes

\item can you see the floor ? \hfill  yes

\item are the walls visible ? \hfill  yes
    \end{compactenum}}
    \end{minipage}}
    &
    \resizebox{6.5cm}{!}{
    \begin{minipage}{2.5in}
    \vspace{-45pt}
    {\tiny
    \begin{compactenum}[\hspace{-5pt}]
\item is the kitchen clean ? \hfill  brown

\item is the kitchen clean ? \hfill  it 's cluttered

\item are there any windows ? \hfill  yes

\item is it daytime ? \hfill  yes

\item are there any people ? \hfill  yes

\item are the appliances stainless steel ? \hfill  i ca n't tell

\item are the people in the kitchen ? \hfill  yes

\item is the kitchen clean ? \hfill  yes

\item are the people male or female ? \hfill  yes

\item is the kitchen clean ? \hfill  yes
    \end{compactenum}}
    \end{minipage}}
    \\[-2pt] \midrule
    \raisebox{0pt}{\includegraphics[height=2.3cm,width=2.7cm]{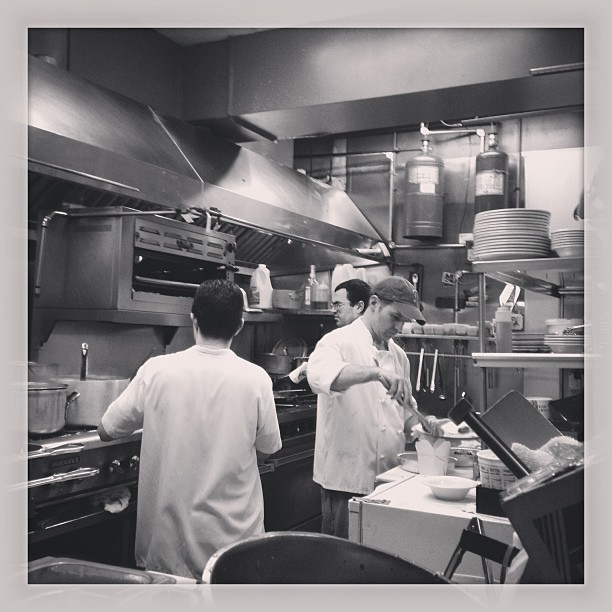}}\newline\vspace{-12pt}
    \begin{minipage}{12cm}{\scriptsize black and white photo of men working in a kitchen}\end{minipage}
    &
    \resizebox{4cm}{!}{
    \begin{minipage}{1.6in}
    \vspace{-55pt}
    {\tiny
    \begin{compactenum}[\hspace{-5pt}]
\item how many men ? \hfill  3

\item how big is the kitchen ? \hfill  small industrial

\item what is the age range ? \hfill  teens to 40s

\item is the kitchen belong in a house or a restaurant ? \hfill  restaurant

\item what are they doing in the kitchen ? \hfill  cooking and mixing food

\item what kind of floor in the kitchen ? \hfill  i ca n't tell

\item is there any decorations in the kitchen ? \hfill  no

\item is the kitchen look clean ? \hfill  no

\item does the men seem to be UNK ? \hfill  they seem UNK    \end{compactenum}}
    \end{minipage}}
    &
    \resizebox{5.3cm}{!}{
    \begin{minipage}{2in}
    \vspace{-61pt}
    {\tiny
    \begin{compactenum}[\hspace{-5pt}]
\item can you see the fridge ? \hfill  the side of it

\item what color is it ? \hfill  the photo is black and white

\item can you see the sink ? \hfill  1 of it

\item can you see the sink ? \hfill  just a tiny bit of it

\item can you see the sink ? \hfill  0

\item can you see the sink ? \hfill  does n't look like it

\item can you see the sink ? \hfill  0

\item can you see the sink ? \hfill  4 of them

\item can you see the fridge ? \hfill  UNK

\item can you see the fridge ? \hfill  not real
    \end{compactenum}}
    \end{minipage}}
    &
    \resizebox{6.5cm}{!}{
    \begin{minipage}{2.5in}
    \vspace{-55pt}
    {\tiny
    \begin{compactenum}[\hspace{-5pt}]
\item how many men are there ? \hfill  looks like 2 men and 1 woman

\item how old do they look ? \hfill  it 's hard to tell , but they look like they are in their 20s

\item how old do they look ? \hfill  looks to be in their 30 's

\item are they wearing aprons ? \hfill  just 1 of them is wearing hat

\item are they wearing hats ? \hfill  it is hard to tell , but i think so

\item can you see the stove ? \hfill  not at all

\item are there any pots on the stove ? \hfill  not that i can see

\item does the kitchen look clean ? \hfill  black and white photo but i think it 's clean

\item can you see any windows ? \hfill  no

\item do you see any windows ? \hfill  ca n't see any windows
    \end{compactenum}}
    \end{minipage}}
    \\[-2pt] \midrule
    \raisebox{0pt}{\includegraphics[height=2.6cm,width=2.7cm]{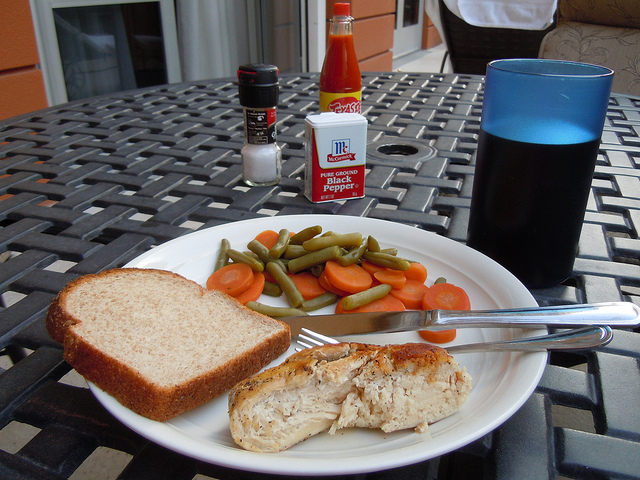}}\newline\vspace{-12pt}
    \begin{minipage}{11cm}{\scriptsize  a half-eaten chicken breast is next to an UNK piece of bread and a small serving of green beans and carrots}\end{minipage}
    &
    \resizebox{4cm}{!}{
    \begin{minipage}{1.5in}
    \vspace{-63pt}
    {\tiny
    \begin{compactenum}[\hspace{-5pt}]
    \item who is eating ? \hfill  is the drink in the cup soda

\item what is the food on ? \hfill  bread and vegetable

\item is it in a plate ? \hfill  what kind of sauces are there in the back

\item i am asking the questions here ? \hfill  yes you are

\item what is the food on ? \hfill  bread , vegetables , chicken and drink

\item are they on a table ? \hfill  yes they are

\item any people ? \hfill  there are n't any people in sight

\item what color is the table ? \hfill  it is dark gray

\item any cutlery ? \hfill  yes there is a butter knife in the plate
    \end{compactenum}}
    \end{minipage}}
    &
    \resizebox{5.3cm}{!}{
    \begin{minipage}{2in}
    \vspace{-67pt}
    {\tiny
    \begin{compactenum}[\hspace{-5pt}]
\item is the photo in color ? \hfill  it is

\item is the photo close up ? \hfill  no

\item is the photo in color ? \hfill  it is

\item do you see any other people ? \hfill  it 's a close up of the man 's hand , so i ca n't tell

\item how old is the man ? \hfill  maybe in his 30s

\item is he wearing glasses ? \hfill  i ca n't see his face

\item what color is his shirt ? \hfill  not sure , the photo is in black and white

\item does he have facial hair ? \hfill  no

\item does he wear glasses ? \hfill  yes
    \end{compactenum}}
    \end{minipage}}
    &
    \resizebox{6.5cm}{!}{
    \begin{minipage}{2.5in}
    \vspace{-75pt}
    {\tiny
    \begin{compactenum}[\hspace{-5pt}]
\item are there any people in the photo ? \hfill  yes

\item is the photo in color ? \hfill  it is

\item is it a professional photo ? \hfill  i do n't think so

\item how many people ? \hfill  0

\item is there a tablecloth on the table ? \hfill  yes

\item what color is the tablecloth ? \hfill  white

\item are there any utensils visible ? \hfill  no

\item what color is the plate ? \hfill  it 's white

\item are there any utensils visible ? \hfill  no

\item are there any utensils visible ? \hfill  no
    \end{compactenum}}
    \end{minipage}}
    \\[-2pt] \midrule

    \end{tabular}}\\[5pt]
    \caption{Selected examples of dialog on v$1.0$ VisDial test split for different SL variants. We observe that variant involving Diverse-\Qbot generates more diverse, image relevant and fluent dialog.}
    \label{table:sl}
    \end{center}
\end{table*}

\begin{table*}[p]
    \vspace{-40pt}
	\renewcommand*{\arraystretch}{0.8}
    \setlength{\tabcolsep}{8px}
    \begin{center}\hspace{-5.5pt}
    \resizebox{2.10\columnwidth}{!}{
    \begin{tabular}{ p{2.85cm}  p{4cm}  p{5.3cm}  p{6.5cm}  }
    \toprule
    {\small Image + Caption} & {\small GT dialog~ \cite{visdial}} & {\small \cite{visdial_rl} dialog} & {\small RL: Diverse-\Qbot-\Abot dialog} \\
    \cmidrule(r){1-1}\cmidrule(lr){2-2}\cmidrule(l){3-3}\cmidrule(l){4-4}
    \raisebox{0pt}{\includegraphics[width=2.7cm]{figures/carrots.jpg}}\newline\vspace{-12pt}
    \begin{minipage}{10cm}{\scriptsize a cutting board with carrots next to a cell phone.}\end{minipage}
    &
    \resizebox{4cm}{!}{
    \begin{minipage}{1.71in}
    \vspace{-55pt}
    {\tiny
    \begin{compactenum}[\hspace{-5pt}]
    \item what color is the cutting board?   \hfill blue
    \item how big is it  \hfill about the size of a regular laptop screen
    \item how many carrots are there ? \hfill 4
    \item are they the orange kind ?  \hfill 2 are orange , 2 are white
    \item what kind of cell phone is it ? \hfill a samsung
    \item can you see any text on the screen ? \hfill a recipe for UNK UNK soup
    \item do you see the room ?  \hfill just the counter
    \item are there people ? \hfill not visible
    \item what color is the counter ?  \hfill white
    \end{compactenum}}
    \end{minipage}}
    &
    \resizebox{5.3cm}{!}{
    \begin{minipage}{2in}
    \vspace{-44pt}
    {\tiny
    \begin{compactenum}[\hspace{-5pt}]
\item is the photo in color ? \hfill  no
\item is the photo close up ? \hfill  no
\item how many people ? \hfill  cows
\item is this a close up ? \hfill  2 of them are
\item what are they wearing ? \hfill  no shirt and jeans
\item what are they wearing ? \hfill  no shirt and jeans
\item do you see other people ? \hfill  sunny 1 in the background
\item what color are the scissors ? \hfill  no handle
\item is this in a kitchen ? \hfill  yes
\item do you see the counter ? \hfill  no
    \end{compactenum}}
    \end{minipage}}
    &
    \resizebox{6.5cm}{!}{
    \begin{minipage}{2.5in}
    \vspace{-45pt}
    {\tiny
    \begin{compactenum}[\hspace{-5pt}]
\item is the photo in color ? \hfill  black and white

\item is there any people ? \hfill  1 hand

\item is this indoors or outdoors ? \hfill  no idea

\item is the person male or female ? \hfill  up close , i ca n't tell

\item can you tell what kind of sandwich it is ? \hfill  no

\item is there a knife ? \hfill  no

\item is there a cutting board ? \hfill  yes

\item are the carrots cut up ? \hfill  daytime

\item is the scene well lit ? \hfill  no

\item are the carrots raw or cooked ? \hfill  short
    \end{compactenum}}
    \end{minipage}}
    \\[-2pt]  \midrule\iftrue
    \raisebox{0pt}{\includegraphics[width=2.7cm]{figures/cows_green.jpg}}\newline\vspace{-12pt}
    \begin{minipage}{10cm}{\scriptsize many brown and black cows standing in the grass}\end{minipage}
    &
    \resizebox{4cm}{!}{
    \begin{minipage}{1.6in}
    \vspace{-45pt}
    {\tiny
    \begin{compactenum}[\hspace{-5pt}]
\item is the field large ? \hfill  yes

\item is the grass green or brown ? \hfill  green

\item can you see a barn ? \hfill  no

\item can you see a fence ? \hfill  no

\item can you see a water supply ? \hfill  yes

\item can you see any people ? \hfill  no

\item can you see trees ? \hfill  yes

\item can you see mountains in the background ? \hfill  no

\item is the grass tall or short ? \hfill  short
\end{compactenum}}
    \end{minipage}}
    &
    \resizebox{5.3cm}{!}{
    \begin{minipage}{2in}
    \vspace{-35pt}
    {\tiny
    \begin{compactenum}[\hspace{-5pt}]
\item can you see any people ? \hfill  nope

\item can you see the sky ? \hfill  a little bit

\item are there any clouds ? \hfill  i ca n't see the sky

\item are there any other animals ? \hfill  not that i can see

\item do you see any trees ? \hfill  lots of trees

\item how many cows are there ? \hfill  ca n't tell , they are far away

\item do you see any buildings ? \hfill  i do not

\item can you see the sky ? \hfill  ca n't see the sky

\item do you see any clouds ? \hfill  ca n't see the sky

\item can you see the sky ? \hfill  ca n't see the sky
 \end{compactenum}}
    \end{minipage}}
    &
    \resizebox{6.5cm}{!}{
    \begin{minipage}{2.6in}
    \vspace{-38pt}
    {\tiny
    \begin{compactenum}[\hspace{-5pt}]
    \item are the cows all the same color ? \hfill  they are different shades of brown

\item do any of the cows have horns ? \hfill  it does n't look like it

\item do you see a fence ? \hfill  there is no fence

\item how many cows are there ? \hfill  4

\item can you see the sky ? \hfill  1 of it

\item what color is the grass ? \hfill  the grass is green

\item can you see the sky ? \hfill  it is not visible

\item does it appear to be sunny ? \hfill  not really

\item can you see the sky ? \hfill  not at all

\item can you see the ground ? \hfill  a little bit
    \end{compactenum}}
    \end{minipage}}
    \\[-2pt] \midrule
    \fi
    \raisebox{0pt}{\includegraphics[width=2.7cm]{figures/elephants.jpg}}\newline\vspace{-12pt}
    \begin{minipage}{10cm}{\scriptsize  many elephants standing in an enclosure near a building}\end{minipage}
    &
    \resizebox{4cm}{!}{
    \begin{minipage}{1.5in}
    \vspace{-50pt}
    {\tiny
    \begin{compactenum}[\hspace{-5pt}]
\item what type of building ? \hfill  may be part of a zoo

\item how many elephants ? \hfill  about 4

\item is it sunny ? \hfill  clear day , yes

\item any trees ? \hfill  yes

\item what color is the building ? \hfill  dark grey

\item are there other animals ? \hfill  no

\item are there any people ? \hfill  yes

\item what are they wearing ? \hfill  shorts and shirts

\item how many ? \hfill  3
    \end{compactenum}}
    \end{minipage}}
    &
    \resizebox{5.3cm}{!}{
    \begin{minipage}{2in}
    \vspace{-45pt}
    {\tiny
    \begin{compactenum}[\hspace{-5pt}]
\item do you see people ? \hfill  2

\item how many elephants ? \hfill  not sure

\item does the elephant have tusks ? \hfill  not that i can see

\item does the elephant have tusks ? \hfill  in the distance , yes

\item does the elephant have tusks ? \hfill  not that i can see

\item any other animals ? \hfill  not that i can see

\item does the elephant have tusks ? \hfill  not that i can see

\item does the elephant have tusks ? \hfill  not that i can see

\item does the elephant have tusks ? \hfill  not that i can see

\item can you see the sky ? \hfill  1 of it
    \end{compactenum}}
    \end{minipage}}
    &
    \resizebox{6.5cm}{!}{
    \begin{minipage}{2.5in}
    \vspace{-48pt}
    {\tiny
    \begin{compactenum}[\hspace{-5pt}]
    \item can you see any people ? \hfill  yes

\item what are the elephants doing ? \hfill  just standing around

\item how many people are there ? \hfill  a lot , too many to count

\item does this look like a zoo ? \hfill  not really

\item how many elephants are there ? \hfill  in the distance there are many

\item does this look like a parade ? \hfill  a little bit

\item does this look like a circus ? \hfill  nope

\item can you see any signs ? \hfill  there is a sign on the street

\item does it look like it is in the usa ? \hfill  i do n't think so

\item can you see the sky ? \hfill  i can
    \end{compactenum}}
    \end{minipage}}
    \\[-2pt] \midrule
    \raisebox{0pt}{\includegraphics[width=2.7cm]{figures/bathroom.jpg}}\newline\vspace{-12pt}
    \begin{minipage}{10cm}{\scriptsize a reflection of the tub is in the bathroom mirror along with the sink , you can see all the room}\end{minipage}
    &
    \resizebox{4cm}{!}{
    \begin{minipage}{1.5in}
    \vspace{-50pt}
    {\tiny
    \begin{compactenum}[\hspace{-5pt}]
\item is the photo in color ? \hfill  yes

\item is there humans in the photo ? \hfill  no

\item is there any animals in the photo ? \hfill  no

\item what color are the walls ? \hfill  tan

\item what color is the sink ? \hfill  white

\item is there any towels ? \hfill  no

\item is there any pictures on the wall ? \hfill  no

\item what color is the tub ? \hfill  white

\item can you see the floor ? \hfill  no
    \end{compactenum}}
    \end{minipage}}
    &
    \resizebox{5.3cm}{!}{
    \begin{minipage}{2in}
    \vspace{-45pt}
    {\tiny
    \begin{compactenum}[\hspace{-5pt}]
    \item how big is the bathroom ? \hfill  ca n't see the whole room

\item how big is the mirror ? \hfill  i can only see a small part of it

\item what color is the sink ? \hfill  there is no sink

\item what color is the sink ? \hfill  black

\item what color is the sink ? \hfill  yes

\item what color are the walls ? \hfill  he is white

\item is there a mirror ? \hfill  yes

\item are there any people ? \hfill  he is alone

\item are there any animals ? \hfill  white cat

\item what color is the sink ? \hfill  yes
    \end{compactenum}}
    \end{minipage}}
    &
    \resizebox{6.5cm}{!}{
    \begin{minipage}{2.5in}
    \vspace{-49pt}
    {\tiny
    \begin{compactenum}[\hspace{-5pt}]
    \item how big is the bathroom ? \hfill  it 's pretty big

\item is it clean ? \hfill  maybe

\item how big is the mirror ? \hfill  i can only see a small part of it

\item what color is the wall ? \hfill  maybe white

\item is there a shower curtain ? \hfill  there is no shower curtain

\item is there a mirror ? \hfill  yes

\item what shape is the mirror ? \hfill  yes , it 's a rectangle

\item what color is the wall ? \hfill  i ca n't see the wall

\item are there any people ? \hfill  he 's the only 1

\item is there a window ? \hfill  i do n't see 1
    \end{compactenum}}
    \end{minipage}}
    \\[-2pt] \midrule
    \raisebox{0pt}{\includegraphics[width=2.7cm]{figures/cows.jpg}}\newline\vspace{-12pt}
    \begin{minipage}{10cm}{\scriptsize a bunch of cows that are in a grassy field}\end{minipage}
    &
    \resizebox{4cm}{!}{
    \begin{minipage}{1.5in}
    \vspace{-40pt}
    {\tiny
    \begin{compactenum}[\hspace{-5pt}]
\item about how many cows ? \hfill  over a dozen

\item are there any calves ? \hfill  no

\item are they eating ? \hfill  yes

\item is it sunny ? \hfill  no

\item is there a fence around them ? \hfill  no

\item do you see any buildings ? \hfill  no

\item mountains in the background ? \hfill  no

\item is there snow on the ground ? \hfill  no

\item any trees ? \hfill  yes
    \end{compactenum}}
    \end{minipage}}
    &
    \resizebox{5.3cm}{!}{
    \begin{minipage}{2in}
    \vspace{-33pt}
    {\tiny
    \begin{compactenum}[\hspace{-5pt}]
    \item is the photo in color ? \hfill  yes

\item is there any people ? \hfill  no

\item how many cows ? \hfill  black and white cows

\item is there a fence ? \hfill  no

\item is there a fence ? \hfill  no

\item is there a fence ? \hfill  no

\item is there a fence ? \hfill  no

\item is there a fence ? \hfill  no

\item is there a fence ? \hfill  no

\item is there a fence ? \hfill  no
    \end{compactenum}}
    \end{minipage}}
    &
    \resizebox{6.5cm}{!}{
    \begin{minipage}{2.5in}
    \vspace{-33pt}
    {\tiny
    \begin{compactenum}[\hspace{-5pt}]
    \item how many cows are there ? \hfill  about 10

\item what color are the cows ? \hfill  black and white

\item are any of them babies ? \hfill  no

\item how many cows are there ? \hfill  about 10

\item are there any people in the photo ? \hfill  no

\item are there any other animals in the photo ? \hfill  no

\item are there any buildings in the photo ? \hfill  no

\item are there any trees ? \hfill  yes

\item is there a fence ? \hfill  no

\item is it sunny out ? \hfill  yes
    \end{compactenum}}
    \end{minipage}}
    \\[-2pt] \midrule
    \raisebox{0pt}{\includegraphics[width=2cm,height=2cm]{figures/kitchen.jpg}}\newline\vspace{-12pt}
    \begin{minipage}{10cm}{\scriptsize the kitchen is UNK with dark wood cabinets and a nearby staircase}\end{minipage}
    &
    \resizebox{4cm}{!}{
    \begin{minipage}{1.53in}
    \vspace{-46pt}
    {\tiny
    \begin{compactenum}[\hspace{-5pt}]
\item what color are the kitchen walls ? \hfill  white

\item is there a window present ? \hfill  yes

\item are there pictures paintings hanging up ? \hfill  no

\item is the stove visible ? \hfill  yes

\item what color is the stove ? \hfill  black and silver

\item are there any people visible ? \hfill  no

\item what color is the nearby staircase ? \hfill  dark brown

\item is the staircase carpeted ? \hfill  no

\item are the handrails of the staircase visible ? \hfill  yes
    \end{compactenum}}
    \end{minipage}}
    &
    \resizebox{5.3cm}{!}{
    \begin{minipage}{2in}
    \vspace{-43pt}
    {\tiny
    \begin{compactenum}[\hspace{-5pt}]
    \item is this in color ? \hfill  no

\item do you see any people ? \hfill  no

\item do you see a stove ? \hfill  standing

\item are there any windows ? \hfill  white windows

\item can you see the floor ? \hfill  metal floor

\item can you see the floor ? \hfill  brown tile

\item can you see the floor ? \hfill  yes

\item is it tiled ? \hfill  no

\item do you see a window ? \hfill  short window

\item are there curtains on the window ? \hfill  yes
    \end{compactenum}}
    \end{minipage}}
    &
    \resizebox{6.5cm}{!}{
    \begin{minipage}{2.5in}
    \vspace{-45pt}
    {\tiny
    \begin{compactenum}[\hspace{-5pt}]
    \item what color are the appliances ? \hfill  black and white

\item is there a window ? \hfill  yes

\item is there a sink ? \hfill  yes

\item can you see a refrigerator ? \hfill  yes

\item are there any people ? \hfill  yes

\item is it daytime ? \hfill  nope

\item is it sunny out ? \hfill  just a little

\item is there a window ? \hfill  brown

\item can you see the floor ? \hfill  no

\item do you see a stove ? \hfill  no
    \end{compactenum}}
    \end{minipage}}
    \\[-2pt] \midrule
    \raisebox{0pt}{\includegraphics[height=2.3cm,width=2.7cm]{figures/black_white_kitchen.jpg}}\newline\vspace{-12pt}
    \begin{minipage}{10cm}{\scriptsize black and white photo of men working in a kitchen}\end{minipage}
    &
    \resizebox{4cm}{!}{
    \begin{minipage}{1.6in}
    \vspace{-57pt}
    {\tiny
    \begin{compactenum}[\hspace{-5pt}]
\item how many men ? \hfill  3

\item how big is the kitchen ? \hfill  small industrial

\item what is the age range ? \hfill  teens to 40s

\item is the kitchen belong in a house or a restaurant ? \hfill  restaurant

\item what are they doing in the kitchen ? \hfill  cooking and mixing food

\item what kind of floor in the kitchen ? \hfill  i ca n't tell

\item is there any decorations in the kitchen ? \hfill  no

\item is the kitchen look clean ? \hfill  no

\item does the men seem to be UNK ? \hfill  they seem UNK    \end{compactenum}}
    \end{minipage}}
    &
    \resizebox{5.3cm}{!}{
    \begin{minipage}{2.2in}
    \vspace{-65pt}
    {\tiny
    \begin{compactenum}[\hspace{-5pt}]
    \item can you see the fridge ? \hfill  yep

\item can you see the fridge ? \hfill  kind of

\item does it have a freezer on it ? \hfill  nope

\item how many men are there ? \hfill  ca n't tell

\item how old do they look ? \hfill  it 's hard to tell , but i 'd say in their 30 's

\item what color are the walls ? \hfill  it 's a black and white photo so i ca n't tell

\item how old are the men ? \hfill  i ca n't see their faces , i do n't know

\item does the kitchen look clean ? \hfill  0

\item how old do the men look ? \hfill  i ca n't see their faces , i do n't know

\item are they wearing aprons ? \hfill  a few are
    \end{compactenum}}
    \end{minipage}}
    &
    \resizebox{6.5cm}{!}{
    \begin{minipage}{2.5in}
    \vspace{-55pt}
    {\tiny
    \begin{compactenum}[\hspace{-5pt}]
    \item can you tell what kind of food they are ? \hfill  i can not tell what kind of food it is

\item does the kitchen look clean ? \hfill  fairly clean

\item do you see any windows ? \hfill  partially in the background

\item do you see any curtains ? \hfill  maybe 1 in the background

\item is it day or night ? \hfill  ca n't tell

\item are the men old or young ? \hfill  ca n't see their faces

\item are they wearing hats ? \hfill  just 1

\item are they wearing glasses ? \hfill  it looks like they are

\item are they smiling ? \hfill  i ca n't see their faces

\item are they smiling ? \hfill  i ca n't see their faces
    \end{compactenum}}
    \end{minipage}}
    \\[-2pt] \midrule
    \raisebox{0pt}{\includegraphics[height=2.6cm,width=2.7cm]{figures/plate_food.jpg}}\newline\vspace{-12pt}
    \begin{minipage}{11cm}{\scriptsize  a half-eaten chicken breast is next to an UNK piece of bread and a small serving of green beans and carrots}\end{minipage}
    &
    \resizebox{4cm}{!}{
    \begin{minipage}{1.5in}
    \vspace{-63pt}
    {\tiny
    \begin{compactenum}[\hspace{-5pt}]
    \item who is eating ? \hfill  is the drink in the cup soda

\item what is the food on ? \hfill  bread and vegetable

\item is it in a plate ? \hfill  what kind of sauces are there in the back

\item i am asking the questions here ? \hfill  yes you are

\item what is the food on ? \hfill  bread , vegetables , chicken and drink

\item are they on a table ? \hfill  yes they are

\item any people ? \hfill  there are n't any people in sight

\item what color is the table ? \hfill  it is dark gray

\item any cutlery ? \hfill  yes there is a butter knife in the plate
    \end{compactenum}}
    \end{minipage}}
    &
    \resizebox{5.3cm}{!}{
    \begin{minipage}{2in}
    \vspace{-78pt}
    {\tiny
    \begin{compactenum}[\hspace{-5pt}]
    \item are there any people ? \hfill  i see part of a person 's hand

\item what color is the plate ? \hfill  black and white

\item are there any utensils ? \hfill  no

\item are there any utensils ? \hfill  no

\item are there any utensils ? \hfill  silver spoon

\item are there any utensils ? \hfill  beige spoon

\item are there any utensils ? \hfill  i see a fork

\item are there any utensils ? \hfill  yes a fork and a knife

\item are there any napkins ? \hfill  0 that i can see

\item are there any napkins ? \hfill  it 's a close up so i ca n't tell
    \end{compactenum}}
    \end{minipage}}
    &
    \resizebox{6.5cm}{!}{
    \begin{minipage}{2.5in}
    \vspace{-60pt}
    {\tiny
    \begin{compactenum}[\hspace{-5pt}]
    \item are there any people in the picture ? \hfill  yes
\item are they inside or outside ? \hfill  outside
\item what gender are the people ? \hfill  i ca n't tell
\item what are the people doing ? \hfill  1 is looking at the camera , the other is looking at the camera
\item can you tell if they are male or female ? \hfill  i ca n't tell
\item can you tell if the person is male or female ? \hfill  i ca n't see their face , i do n't know
\item can you tell if they are male or female ? \hfill  i ca n't tell
\item can you tell what kind of sandwich it is ? \hfill  it has some meat on it , but i ca n't tell what it is

\item is it on a plate ? \hfill  no

\item what color is the handle of the knife ? \hfill  silver
    \end{compactenum}}
    \end{minipage}}
    \\[-2pt] \midrule

    \end{tabular}}\\[5pt]
    \caption{Selected examples of dialog on v$1.0$ VisDial test split for different RL variants. We observe that variant involving Diverse-\Qbot generates more diverse, image relevant and fluent dialog.}
    \label{table:rl}
    \end{center}
\end{table*}

\end{document}